\documentclass[11pt]{article}

\usepackage[final]{acl}

\usepackage{times}
\usepackage{latexsym}

\usepackage[T1]{fontenc}
\usepackage[utf8]{inputenc}

\usepackage{microtype}

\usepackage{inconsolata}

\usepackage{graphicx}
\usepackage{booktabs}
\usepackage{multirow}
\usepackage{subcaption}
\usepackage{amsmath}
\usepackage{float}
\usepackage{bbding}

\title{Can We Still Trace L1 Signals? \\ Investigating the Resilience of Native Language Signals in the LLM Era}

\author{
Nabelanita Utami \quad Ryohei Sasano \\
Graduate School of Informatics, Nagoya University \\
\texttt{utami.nabelanita.v7@s.mail.nagoya-u.ac.jp, sasano@i.nagoya-u.ac.jp}
}

\begin{document}
\maketitle
\begin{abstract}
The widespread use of LLM-based writing assistance has raised an interesting question about the homogenization of English.
As LLMs tend to revise texts 
toward mainstream English conventions reflected in their training data, the subtle fingerprints that reflect an author's native language (L1) may be gradually disappearing. 
This study investigates this phenomenon by analyzing native language identification (NLI) performance on academic abstracts.
To this end, we construct two NLI datasets of academic abstracts extracted from arXiv and the ACL Anthology that covers eight native language groups across three time periods: pre-neural network (NN), pre-LLM, and post-LLM.
We then evaluate NLI performance for each era using NLI classifiers obtained by fine-tuning LLMs.
The results reveal a consistent decline in NLI performance over time.
Notably, however, the decline is more pronounced from the pre-NN era to the pre-LLM era than from the pre-LLM era to the post-LLM era.
This suggests that although academic English appears to have become increasingly homogenized in the LLM era, this homogenization did not suddenly emerge with the advent of LLMs; rather, it has progressed gradually since the emergence of neural approaches to language processing.
Furthermore, a rewriting experiment using recent LLMs shows a larger NLI performance drop than the progression across eras alone, suggesting that increased LLM use may lead to further homogenization in the future.

\end{abstract}

\section{Introduction}
\label{sec:introduction}

The landscape of academic writing has changed over the past decade due to the development of writing assistance tools. Researchers have moved from using simple dictionary lookups to neural machine translation (NMT), and more recently to large language models (LLMs) that can paraphrase entire sections. Unlike earlier tools, LLMs tend to produce output that leans toward American English conventions that were influenced by their training data and reinforcement learning from feedback \cite{naturebias2025}. Following the release of ChatGPT in 2022, the use of LLMs in academic writing has shown a 17.5\% increase in computer science papers published on arXiv by 2024 \cite{liang2024mappingincreasingusellms}. This raises a question: as writing tools become more capable, are the stylistic traces that reflect an author's native language gradually disappearing?

\begin{figure}[t]
    \centering
    \includegraphics[width=\columnwidth]{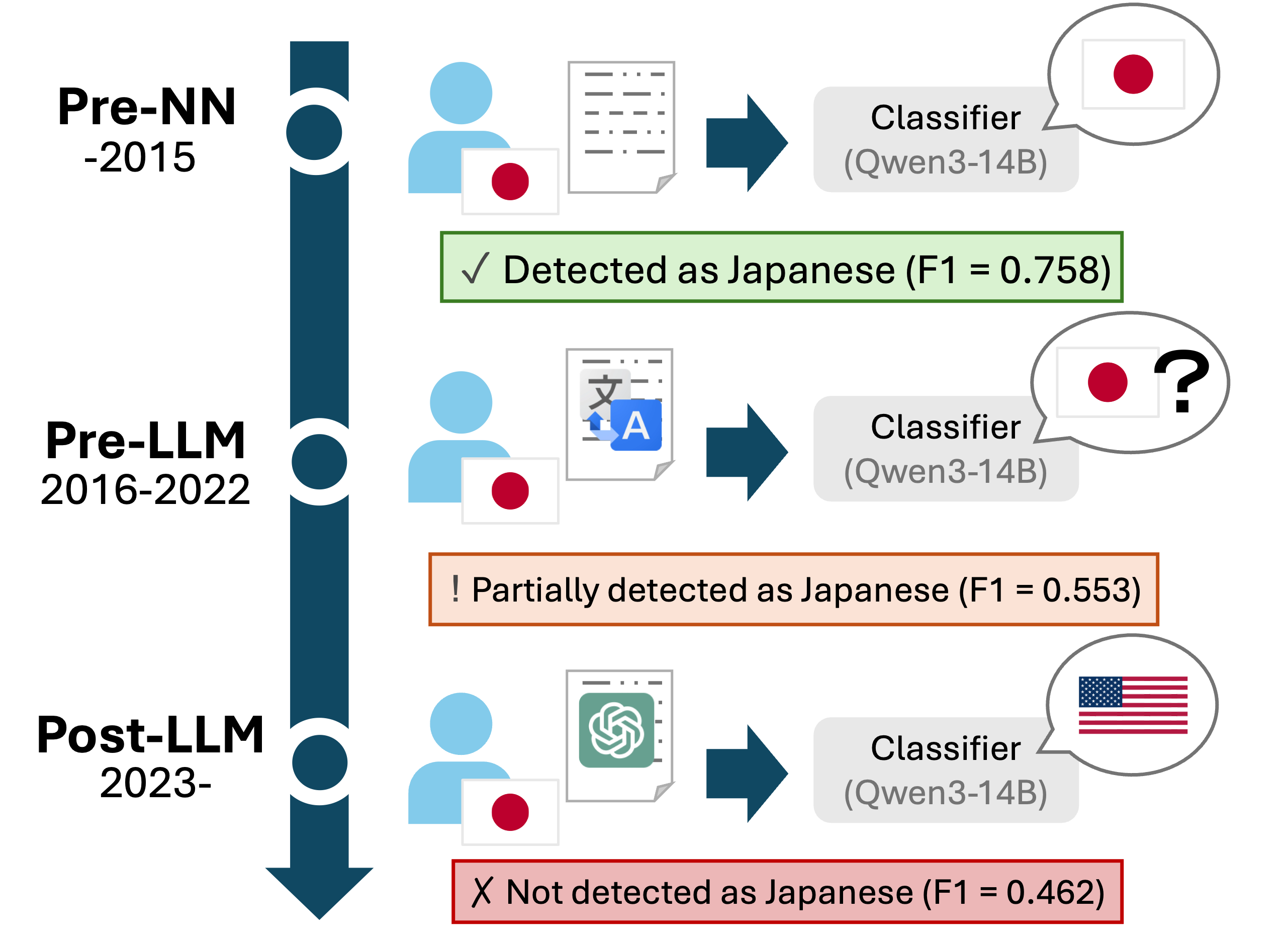}
    \caption{Illustration of the homogenization hypothesis. The figure shows an example based on an abstract written by a Japanese author. In the pre-NN era, the classifier can readily identify the author's L1 as Japanese, but detection becomes increasingly difficult in later eras. The F1-scores shown in the figure are actual scores obtained with a Qwen3-14B-based NLI classifier.}
     \label{fig:intro_figure}
\end{figure}

This is not an entirely new phenomenon. Research on the ``translationese'' effect has shown that translated text often becomes more simplified and standardized \cite{volansky2015translationese}. This occurrence suggests that writing assisted by automated tools may gradually converge toward a dominant style. With the rise of neural machine translation (NMT) around 2016, non-native researchers gained access to translation with higher-fluency. Recent studies have also documented shifts in vocabulary patterns in LLM-era writing, such as increased frequency of certain words \cite{juzek-ward-2025-chatgpt, kobak2024delving}. Writing homogenization may therefore be a gradual process that started before LLMs and has become more noticeable in recent years.

Native language identification (NLI) is the task of predicting an author’s native language (L1) based on their writing. It was first formalized as a text classification problem by \citet{koppel2005determining}. Historically, NLI has focused on learner corpora (e.g., TOEFL essays) that rely on explicit grammatical errors to detect L1 interference signals \cite{tetreault-etal-2013-report3}. Applying NLI to academic writing presents a unique challenge since researchers are typically highly proficient in English. The ``fingerprints'' of native language are not obvious errors in this domain, but are subtle preferences often tied to specific patterns of language use. In high-proficiency academic writing, this often manifests as ``rhetorical transfer,'' which refers to certain structures due to linguistic background \cite{kaplan1966cultural}.

In this work, we investigate the impact of writing assistance tools on the diversity of academic texts by analyzing NLI performance across three distinct technological eras: pre-neural network (NN), pre-LLM, and post-LLM. We hypothesize that L1 stylistic signals have been weakening gradually over time as writing tools improve, rather than declining suddenly with the introduction of LLMs. Figure~\ref{fig:intro_figure} illustrates this hypothesis using a Japanese-authored abstract as an example. Our contributions are as follows:
\begin{itemize}

    \item We construct two datasets specifically for high-fluency academic writing, extracted from papers published in arXiv and the ACL Anthology, mapped to author demographics across three historical eras.\footnote{Datasets are publicly available at \url{https://github.com/nabelanita/academic-l1-signals}.}
    \item We analyze NLI performance across three eras and find that the largest drop occurs between the pre-NN and pre-LLM eras suggesting that homogenization was already underway during the rise of NMT tools.
    \item We conduct a rewriting experiment showing that LLM-based rewriting reduces L1 signals across all eras, including texts with less characteristics. This suggests that LLMs amplify an existing trend rather than causing it.
\end{itemize}

\section{Related Work}
\label{sec:related_work}
\subsection{Native Language Identification}
Native language identification is the task of predicting an author's native language (L1) based solely on their writing in a second language (L2). It was first formalized as a text classification problem by \citet{koppel2005determining}, who demonstrated that an unknown author's L1 could be predicted from their English writing style with high accuracy. Since then, NLI has grown into an established research area with shared tasks \cite{tetreault-etal-2013-report3} and a comprehensive body of literature recently surveyed by \citet{goswami-etal-2024-native}.

Historically, native language identification has focused on learner corpora, such as collections of writing by language students like TOEFL essays. It relies on explicit grammatical errors as signals of L1 interference \cite{tetreault-etal-2013-report3}. In this setting, the task is relatively tractable because learner errors are frequent and easily identified. However, applying native language identification to academic writing presents a different challenge. \citet{stehwien-pado-2015} constructed a novel corpus of scientific papers fron ACL Anthology labeled by author L1. They showed that NLI models trained on learner corpora generalize poorly to scientific writing. This highlights that academic text constitutes a distinct and more difficult text type for NLI. Researchers publishing in peer-reviewed venues tend to be relatively proficient in English, and their writing has usually undergone strict editing. The fingerprints of native language are therefore not obvious grammatical errors, but rather subtle stylistic preferences related to the author's native language. In writings with high fluency, influences of L1 often turn into rhetorical transfer. It is a preference for certain structures and patterns due to cultural influence \cite{kaplan1966cultural}. Prior work has shown that such transfer patterns are detectable even in fluent academic writing through features such as syntactic complexity, discourse and lexical choices \cite{jarvis-2012-approaching, tsvetkov2013identifying, malmasi2018native}.

More recently, \citet{nicholls-alperin-2025-cross} showed that open-source LLMs are effective for NLI across diverse genres. Additionally, they find document length being an important factor for performance. \citet{goswami-etal-2025-multilingual} evaluated LLMs on NLI beyond English, extending the task to multilingual corpora and showing that LLM-based approaches can also be effective in non-English settings. Our work differs from these studies in that we use fine-tuned LLMs not to advance NLI performance, but as a measurement tool to track how L1 signals in academic writing change over time.

\subsection{L1 Transfer and Writing Assistance Tools}
The relationship between L1 interference and writing assistance has a longer history than LLMs. Early tools such as spell checkers and grammar correctors could already suppress some surface-level L1 errors. With the rise of neural machine translation (NMT), non-native researchers gained access to tools capable of translating entire sentences with near-fluent output. This produced a phenomenon analogous to translationese. It refers to texts that are grammatically correct but stylistically simplified. The text lacks the variation of native writing \cite{volansky2015translationese}. Research on translationese suggests that text produced through translation tends to be similar to the dominant style in the system's training data. This is relevant to our work because LLM-assisted writing can be seen as a similar process,  but for revision and paraphrasing.

\subsection{LLM Use in Scientific Writings}
The release of large language models to the general public, most notably GPT-3.5 through ChatGPT in November 2022 and GPT-4 in March 2023, has fundamentally changed how people write, extending beyond academia to the general society \cite{liang2025widespreadadoptionlargelanguage}. The adoption of these tools in scientific publishing has been rapid and measurable. \citet{liang2024mappingincreasingusellms} conducted the first large-scale analysis of LLM use in academic writing, examining 950,965 papers published on arXiv, bioRxiv, and Nature portfolio journals between January 2020 and February 2024. Their findings revealed a 17.5\% increase in LLM-modified content in computer science papers by early 2024, subsequently updated to 22\% in a follow-up analysis \cite{liang2025nature}.

Similar studies have also examined word patterns linked to LLM use. \citet{kobak2024delving} showed that words such as ``delve,'' ``underscore,'' and ``intricate'' increased by over 1,000\% in frequency between 2022 and 2024. \citet{juzek-ward-2025-chatgpt} argue that this overuse is a result of reinforcement learning from human feedback (RLHF), which steers models toward an output rated as polished by human annotators \cite{ouyang2022training, lambert2026rlhf}. As a result, the model tends to produce a more uniform style of formal academic English.

This is especially relevant for non-native speakers. \citet{cong2024divergent} found significantly higher rates of LLM-assisted writing among non-native English speakers compared to native speakers. This is consistent with LLMs being perceived as useful for overcoming the language barrier in academic publishing. Furthermore, LLM training corpora are heavily biased toward English text, particularly toward Standard American English \cite{naturebias2025}. This suggests that when a non-native speaker uses an LLM to revise their writing, the model steers the text toward American English norms.

\section{Dataset}
\label{sec:dataset}
\subsection{Dataset Construction: Semi-Automated Labeling Framework}
\label{sec:dataset_construction}
Given the lack of large-scale scientific paper datasets labeled with author native languages, we developed a semi-automated framework to produce high-confidence labels. Our pipeline utilizes an LLM-augmented labeling strategy that combines metadata with LLM-predicted name origins.
Specifically, our labeling workflow proceeds in three stages.
The first two stages involve estimating the author’s country of origin, followed by a third stage in which the estimated country is mapped to a corresponding language label.

% \begin{figure*}[t]
%     \centering
%     \includegraphics[width=0.85\textwidth]{figures/labeling_process.png}
%     \caption{The Labeling Pipeline. We verify author labels by intersecting LLM-based name predictions (Top-2) with institutional affiliations. Authors with mixed Anglosphere affiliations are strictly filtered out to ensure distinct L1 signals.}
%     \label{fig:labeling_pipeline}
% \end{figure*}

\paragraph{1) Author-Level Verification.}
For each paper, we retrieved author names and institutional affiliations via the OpenAlex API \cite{priem2022openalexfullyopenindexscholarly}. 
To resolve ambiguities caused by researcher migrations, we prompted Qwen3-8B \cite{yang2025qwen3technicalreport} to predict the top-2 most likely countries of origin based solely on the author's name, then intersected this with the author's affiliation country. We assign a label only if the affiliation appears within the model's top-2 candidates. To minimize English immersion effects, we excluded non-native candidates with dual affiliations at English-speaking institutions. While this logic minimizes labeling errors, it cannot account for complex linguistic backgrounds (e.g., a researcher with a Chinese name raised in the United States), so our labels represent high-probability estimates rather than ground truth.
% To resolve ambiguities caused by researcher migrations, we employed a cross-verification step:
% \begin{itemize}
%     \item We prompted Qwen3-8B \cite{yang2025qwen3technicalreport} to predict the top-2 most likely countries of origin based solely on the author's name. If certain, the model is allowed to output the same language twice.
%     \item We intersected this prediction with the author's affiliation country. An author is assigned a country label \textit{if and only if} their affiliation appears within the LLM's top-2 candidate list.
%     \item Anglosphere exclusion: To minimize the influence of English immersion on L1 signals, we strictly excluded any non-native candidate who held a dual affiliation with an institution in an English-speaking country (e.g., an author affiliated with both \textit{Tsinghua University} and \textit{Harvard University}). We assume such authors possess high fluency that may conceal their native linguistic fingerprints.
% \end{itemize}
% While our strict logic minimizes errors, it cannot account for complex linguistic backgrounds (e.g., a researcher with a Chinese name who was born and raised in the United States). Consequently, our ``native'' labels serve as a high-probability rather than a ground truth.

\paragraph{2) Paper-Level Consensus.}
To ensure the text has a consistent native language signal, we require strict background coherence across co-authors. We restricted the dataset to papers with five or fewer authors. A final L1 label is assigned to the paper only if the key authors (first, second, and last) share the same verified country label. 
%Additionally, this country must account for a super-majority ($>66\%$) of all authors.

\paragraph{3) Language Mapping.}
Finally, the verified country labels were mapped to their primary official languages (e.g., `US` $\rightarrow$ \texttt{english\_american}, `CN` $\rightarrow$ \texttt{chinese}) using information obtained from Wikidata \cite{wikidata} references.
In this study, to avoid ambiguity in the mapping process, we restrict language mapping to papers assigned a country label that has a primary official language and for which sufficient experimental data can be collected.
The languages used in this study are described in Section \ref{sec:data_statistics}.

\subsection{Training and Evaluation Data Curation}
\label{sec:data_curation}
We conducted our data construction pipeline a total of two times and produced two separate datasets for different purposes. 

\paragraph{1) Training Data (arXiv).} To train a general classifier, we applied the framework described in Section~\ref{sec:dataset_construction} to the arXiv dataset \cite{clement2019usearxivdataset}. From the filtered results, we sampled a balanced subset across target languages to prevent class imbalance. To prevent the model from overfitting to the writing style of a specific time era, we applied a strict sampling cap per publication year. This number varies across languages due to differences in sample availability.

\paragraph{2) Evaluation Data (ACL Anthology).} To test our hypothesis regarding writing homogenization, we compiled a dataset from the ACL Anthology corpus \cite{acl_anthology_corpus} and divided it into three technological eras: pre-NN ($\leq$2015), pre-LLM (2016--2022), and post-LLM (2023--2025). We focus on two major milestones in publicly available machine translation: the release of Google Neural Machine Translation \cite{wu2016googlesneuralmachinetranslation} in 2016 and GPT-4 \cite{openai2024gpt4technicalreport} in 2023. Based on these milestones and the number of papers available for collection, we define three eras: up to 2015, 2016--2022, and 2023--2025. 
\subsection{Data Statistics}
\label{sec:data_statistics}

We focused on eight target languages for which sufficient data could be collected across all three eras: American English, British English, French, German, Italian, Chinese, Japanese, and Korean. For the training set, which is derived from arXiv, we selected a balanced subset of 1,600 samples (200 per language) spanning the years 1999--2021.

For the evaluation set, we select 50 papers for each combination of three eras and eight languages, resulting in a total of 1,200 papers. Since most of these papers have been published through a peer-review process, they exhibit a standardized level of writing quality. This allows our model to capture stylistic characteristics specific to authors' L1, rather than simple grammatical errors. For combinations with fewer than 50 available papers, such as Korean in the pre-NN era, we address this by duplicating a subset of the collected papers. We manually verified this dataset and found that almost all instances were assigned correct language labels. To ensure no overlap between the training and evaluation sets, we cross-checked the arXiv training set against the ACL Anthology evaluation set and found only a single duplicate instance across a combined total of 2,800 papers, which we consider negligible.

\subsection{Linguistic Feature Analysis}
We conducted a linguistic feature analysis on our evaluation dataset across all three periods to examine whether writing style changes are linguistically observable. We selected 9 features tied to L1 transfer errors documented in \citet{swansmith2001learners}. This allows us to track changes more clearly than with a larger feature set. The feature selection process and the final set of features are detailed in Appendix~\ref{app:chosen_features}.

Features were extracted from each abstract using spaCy \cite{Honnibal2020Spacy} and computed as normalized ratios (e.g., per sentence or per total words) to control for variation in abstract length across eras. In Appendix~\ref{app:feature_heatmap}, we present the mean values of the final 9 features per language group across the three eras, with colors normalized per feature across all eras. In the pre-NN era, the non-native language groups exhibit different feature profiles. This reflects L1-specific transfer tendencies documented in prior work \cite{swansmith2001learners}. As an example, East Asian language groups (Japanese, Chinese, Korean) show a lower article usage frequency. This is consistent with the lack of grammatical articles in their grammar. European language groups (French, German, Italian) tend to display higher subordinate clause ratios. This follows the more complex structures in their native languages. As we move into the pre-LLM and post-LLM eras, these differences begin to narrow, specifically closer to American English. Notably, despite being a native English variety, British English also shows a significant stylistic shift. 

\begin{figure}[!t]
    \centering
    \includegraphics[width=0.95\linewidth]{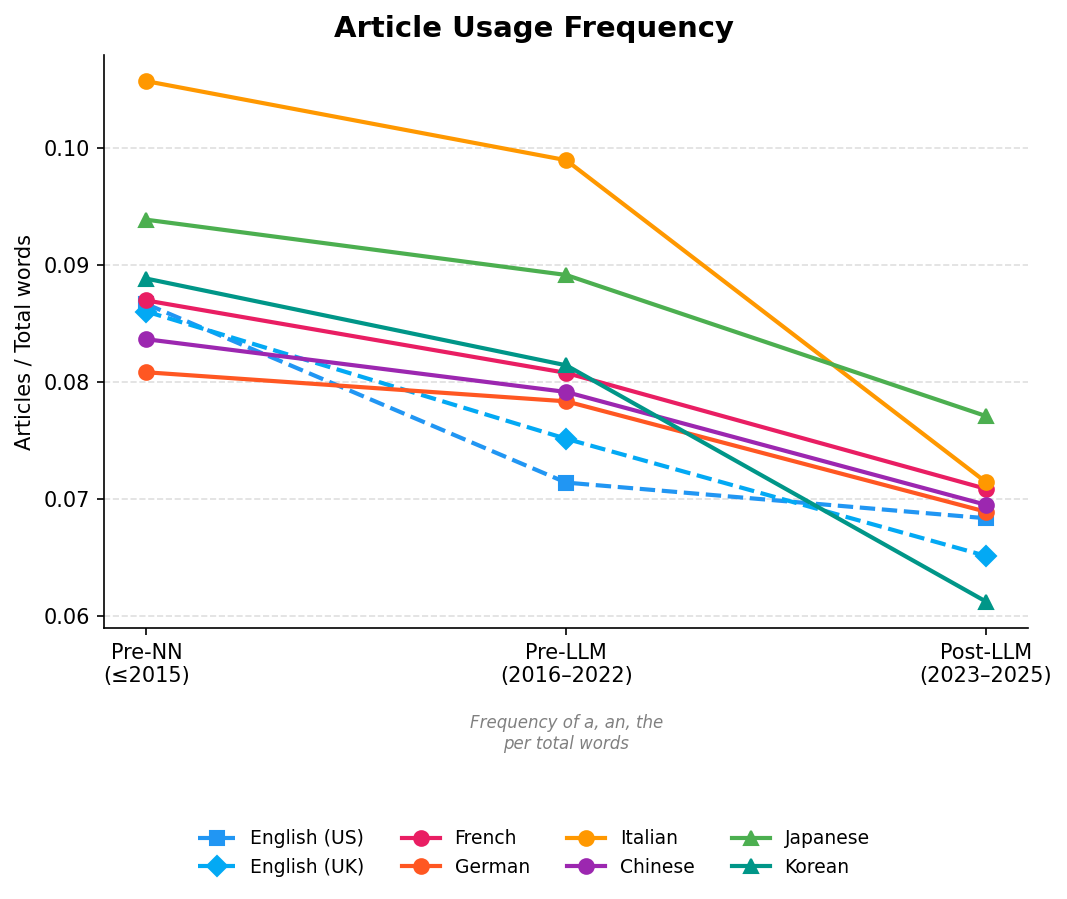}
    \vspace{0.2em}
    \includegraphics[width=0.95\linewidth]{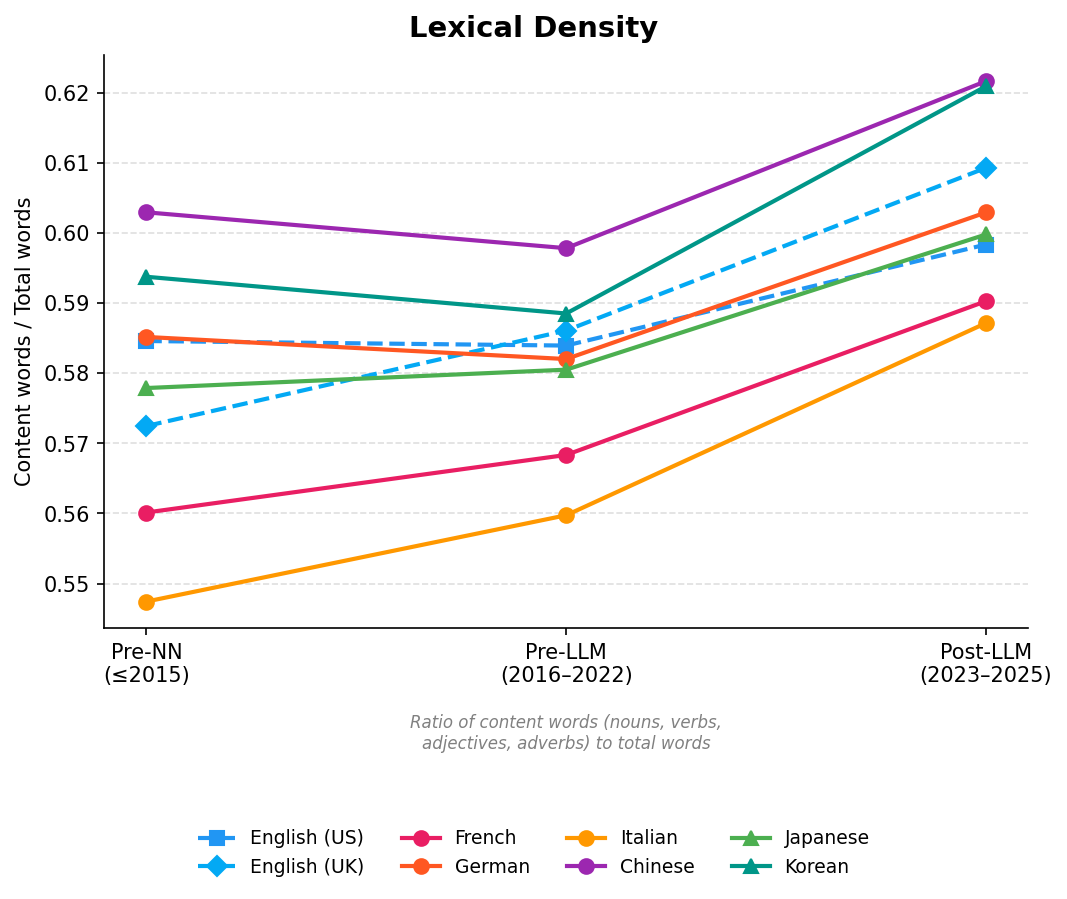}
    \vspace{0.2em}
    \includegraphics[width=0.95\linewidth]{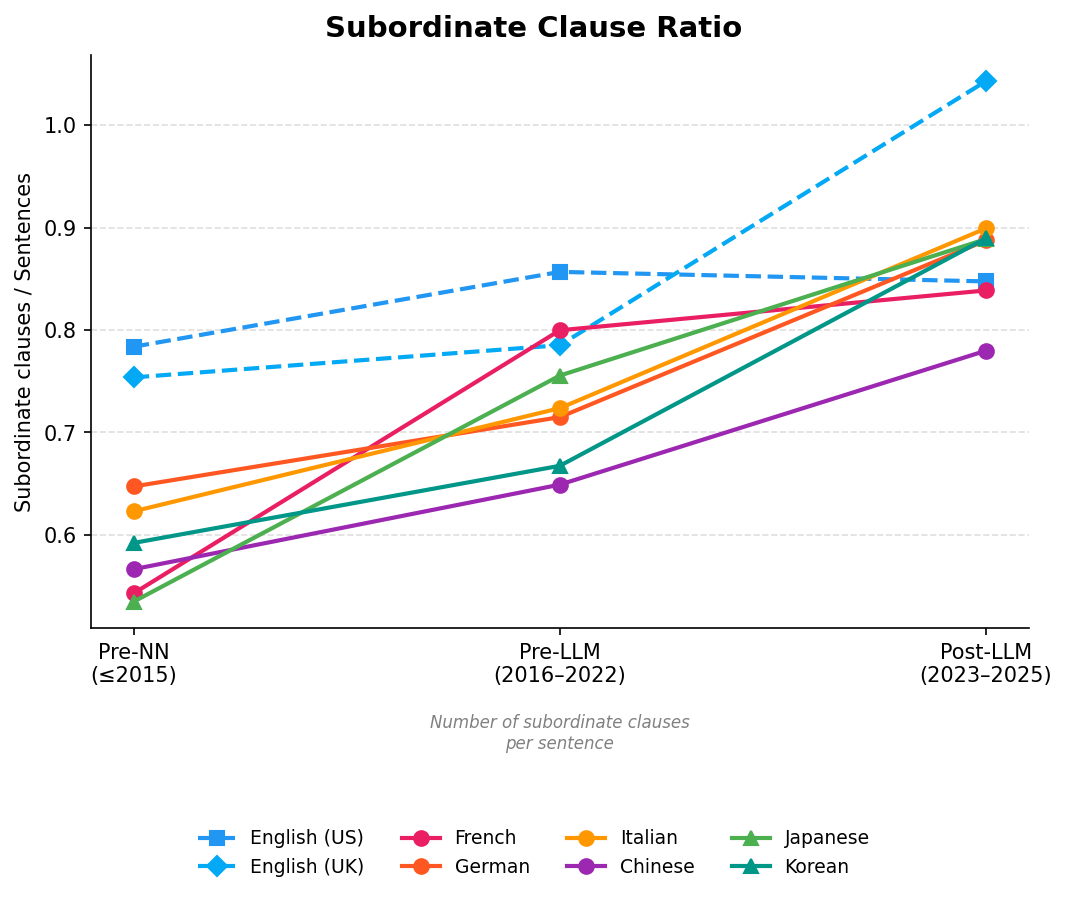}
    \caption{Comparison of shifts across periods observed in three features: article usage frequency, subordinate clause ratio, and lexical density}
    \label{fig:feature_plot_comparison}
\end{figure}
Additionally, Figure~\ref{fig:feature_plot_comparison} shows the three features with the most pronounced changes across periods. A consistent pattern across all three features is the narrowing of this spread in the post-LLM era. This suggests that the stylistic distances between language groups are decreasing over time. Lexical density rises uniformly across all groups in the post-LLM era, likely affected by the tendency of LLM-assisted writing to favor nominalized phrasing. Subordinate clause ratio also increases drastically. Previously low-ratio groups (e.g. Chinese and Korean) are observed catching up to European levels. Article usage frequency shows a downward convergence. The high-usage European groups drop sharply in the post-LLM era, moving toward the lower values which were previously associated with East Asian writers. This suggests a regression toward a neutral LLM-preferred register rather than a native-English one. 

\section{Native Language Identification}
\label{sec:nli_models}

\subsection{Methodology}
\label{sec:nli_methodology}
We introduce two NLI models: one based on few-shot prompting and the other based on fine-tuning. These models are applied to evaluation data from three eras. A relative performance drop on post-LLM-era data suggests that the advent of LLMs has reduced the presence of L1-specific stylistic traces in the English of non-native speakers.

\paragraph{Few-Shot Prompting.}
We prompted the models for classifying native languages of authors by restricting the output to a closed set of language labels. Since peer-reviewed academic text is highly fluent, standard models frequently default to predicting ``native English.'' As a solution, our system prompt explicitly directs the model to identify subtle L1-interference patterns and to avoid assigning English labels without any strong evidence. The input to the model consists of the paper title and abstract, formatted within a specific template. See Appendix~\ref{app:appendix_fewshot_prompt} for the full system prompt.

\paragraph{Fine-Tuning.}
We fine-tuned two open-weights models to evaluate their capability in detecting subtle stylistic fingerprints: Qwen3-14B \cite{yang2025qwen3technicalreport} and Gemma-3-12B-it \cite{gemmateam2025gemma3technicalreport}. We employed QLoRA \cite{qlora} with 4-bit quantization and low-rank adapters (LoRA) \cite{hu2021loralowrankadaptationlarge}. Fine-tuning was performed on a single NVIDIA A100 (80GB) GPU. The training was performed on the balanced arXiv dataset described in Section~\ref{sec:data_curation}, using a maximum sequence length of 1024 tokens. Hyperparameters and the fine-tuning prompt are detailed in Appendices~\ref{app:appendix_hyperpara} and~\ref{app:appendix_prompt_tuning} respectively.

\subsection{Experiment and Results}
\label{sec:nli_experiment}
We evaluated our models on the ACL Anthology evaluation set described in Section~\ref{sec:data_curation}. %This set consists of balanced subsets for three eras: pre-NN ($\le$2015), pre-LLM (2016--2022), and post-LLM (2023--2025).
We first evaluated the base models in a few-shot setting (one example per language, per era) without fine-tuning, then evaluated our fine-tuned variants. We report both accuracy and F1-score. Given that the test sets are balanced, these metrics provide a direct measure of the model's ability to differentiate L1 signals.

Table \ref{tab:main_results} presents the accuracy and F1-score across all eight language classes for the three eras. Both models exhibit the same trend. The highest scores are obtained on pre-NN era data, followed by pre-LLM-era data, with the lowest on post-LLM-era data. This is consistent with our hypothesis that L1 traces in academic writing have weakened over time as writing tools have improved.

\begin{table*}[t]
    \centering
    \small
    % --- TABLE 1: Main Results ---
    \renewcommand{\arraystretch}{0.85}
    \begin{tabular}{ll|ccc|ccc}
    \specialrule{1.5pt}{1pt}{2pt}
    % \toprule
    %& \textbf{Model} 
    & & \multicolumn{3}{c|}{\textbf{Qwen3-14B}} & \multicolumn{3}{c}{\textbf{Gemma-3-12B-it}} \\
    Models & Metric & pre-NN & pre-LLM & post-LLM & pre-NN & pre-LLM & post-LLM \\ \midrule
    %\multicolumn{7}{c}{Few-Shot} \\ \midrule
    \multirow{2}{*}{Few-shot} & Accuracy & \textbf{0.378} & 0.181 & 0.145 & \textbf{0.304} & 0.258 & 0.191 \\
    & F1-score & \textbf{0.393} & 0.137 & 0.067 & \textbf{0.304} & 0.222 & 0.111 \\ \midrule
    %\multicolumn{7}{c}{Fine-tuned} \\ \midrule
    \multirow{2}{*}{Fine-tuned} & Accuracy & \textbf{0.728} & 0.650 & 0.633 & \textbf{0.718} & 0.628 & 0.590 \\
    & F1-score & \textbf{0.726} & 0.637 & 0.620 & \textbf{0.715} & 0.614 & 0.598 \\
    \specialrule{1.5pt}{1pt}{2pt}
    \end{tabular}
    \caption{Native language identification performance (accuracy and F1-score) across three eras.}
    %\caption{Native language identification performance (accuracy and F1-score) across three eras using few-shot prompting and fine-tuning.}
    \label{tab:main_results}
\end{table*}

% First, the few-shot prompt engineering performed poorly, with Qwen3 achieving only 37.8\% accuracy in the pre-NN era and dropping to 14.5\% in post-LLM. 
For the few-shot setting, Qwen3-14B accuracy drops from 0.378 (pre-NN) to 0.145 (post-LLM). Both models show a strong bias toward American English and often collapse into a single-class prediction. Confusion matrices are provided in Appendix~\ref{sec:app_fewshot_results}.

Our fine-tuned models perform significantly better. They achieved over 0.700 accuracy in the pre-NN era. Performance degrades consistently over time, from 0.728 to 0.633 for Qwen3 and from 0.718 to 0.590 for Gemma 3. Detailed results can be seen in Appendix \ref{app:ft_full_result}.

Additionally, we performed Fisher's exact test on the accuracy differences in Table \ref{tab:main_results}. For the fine-tuned models, Fisher’s exact test with a significance level of $\alpha = 0.05$ indicated that the differences in accuracy between the pre-NN era and both later eras are statistically significant for both models (pre-NN vs. pre-LLM: $p = 0.0218$ for Qwen3, $p = 0.0083$ for
Gemma~3; pre-NN vs. post-LLM: $p = 0.0049$ for Qwen3, $p = 0.0002$ for Gemma-3). Notably, the difference between pre-LLM and post-LLM is not significant for either model ($p = 0.6583$ and $p = 0.3105$ respectively). This indicates that the biggest shift in writing style began around 2016 when NMT tools became widely used. The shift was not limited to the LLM era. Full significance test results, including the few-shot setting, are provided in Appendix~\ref{sec:appendix_stats_testing}.

One possible concern is that the lower performance in the post-LLM era is caused by differences between the training and evaluation periods. In other words, it may not necessarily reflect genuine homogenization. However, we note that performance already declines in the pre-LLM era (2016-2022) that partially overlaps with our training data period. This suggests that the decline reflects a genuine change in writing style over time.

\subsection{Discussion}
\label{sec:nli_discussion}
To better understand this performance decline, we looked at class-specific performance (Table~\ref{tab:detailed_metrics}). The results suggest that homogenization is real but not uniform. Two languages, Chinese and French, do not follow the general declining trend, while the rest largely do. 

\begin{table*}
    \centering
    \small
    % --- TABLE 2: Detailed Language Metrics ---
    \begin{tabular}{l|ccc|ccc}
        \specialrule{1.5pt}{1pt}{2pt}
        \textbf{Language} & \multicolumn{3}{c|}{\textbf{Qwen3-14B (fine-tuned)}} & \multicolumn{3}{c}{\textbf{Gemma-3-12B-it (fine-tuned)}} \\
         & pre-NN & pre-LLM & post-LLM & pre-NN & pre-LLM & post-LLM \\ \midrule
        en-US & \textbf{0.648} & 0.574 & 0.593 & \textbf{0.679} & 0.522 & 0.576 \\
        en-GB & \textbf{0.602} & 0.438 & 0.406 & \textbf{0.565} & 0.435 & 0.235 \\
        fr      & 0.703 & \textbf{0.720} & 0.690 & 0.667 & 0.654 & \textbf{0.723} \\
        de      & \textbf{0.686} & 0.604 & 0.612 & \textbf{0.694} & 0.559 & 0.581 \\
        it     & \textbf{0.752} & 0.732 & 0.703 & \textbf{0.788} & 0.739 & 0.688 \\
        zh     & \textbf{0.876} & 0.815 & 0.869 & 0.812 & 0.737 & \textbf{0.885} \\
        ja    & \textbf{0.758} & 0.553 & 0.462 & \textbf{0.757} & 0.615 & 0.508 \\
        ko      & \textbf{0.784} & 0.657 & 0.628 & \textbf{0.761} & 0.651 & 0.590 \\
        \hline
        \textbf{Avg.} & \textbf{0.726} & 0.637 & 0.620 & \textbf{0.715} & 0.614 & 0.598 \\
        \specialrule{1.5pt}{1pt}{2pt}
    \end{tabular}
    \caption{Per-language F1-scores for native language identification across three eras using fine-tuned models.
    }
    \label{tab:detailed_metrics}
\end{table*}

The clearest declines are observed in British English, Japanese, and Korean. For Gemma 3, British English detection dropped from 0.565 (pre-NN) to just 0.235 (post-LLM), with misclassified papers largely predicted as American English. This is consistent with US English dominance in LLM training corpora. Japanese and Korean follow a similar pattern. Japanese F1 dropped from $\approx$0.758 to $\approx$0.462 (Qwen3), and Korean from $\approx$0.784 to $\approx$0.628. Both languages are structurally distant from English, and we suspect writing assistance tools are increasingly correcting characteristic transfer errors, such as SOV-influenced word order and topic-prominent constructions. This effectively removes the features that once made these groups detectable.

Chinese shows the most notable resistance. Its F1-score stays stable or increases across eras, reaching 0.885 with Gemma-3 in the post-LLM era. One possible reason is the split between Western and Chinese AI ecosystems. Chinese researchers tend to rely on domestic models (e.g., Qwen, DeepSeek, GLM) due to restrictions on Western APIs. Additionally, the bilingual nature of these models may normalize writing differently from English-dominant ones~\cite{glm, deepseekai2025deepseekv3technicalreport}. As with the French case below, we do not have direct evidence of which specific tools Chinese-affiliated authors used, so this explanation remains speculative and warrants further investigation.

French, on the other hand, shows mixed trends across models. It shows a slight drop with Qwen3 but an increase with Gemma 3. One possible, speculative explanation relates to the broader French NLP ecosystem, which includes French-language and multilingual models such as CamemBERT \cite{camembert} and Mistral \cite{jiang2023mistral7b}. We do not have evidence that these specific models were used as writing assistants by the authors in our dataset, but their existence suggests that the writing assistance ecosystem available to French researchers is somewhat less English-centric than for other non-native groups, which could plausibly relate to more preserved L1 features. Additionally, French shares more structural features with English than East Asian languages do \cite{swansmith2001learners}, which may independently make French L1 patterns more resilient to normalization by writing assistance tools. We acknowledge that both explanations remain speculative, and a more thorough investigation is left for future work.

\section{LLM Rewriting Experiment}
\label{sec:rewrite_experiment}
To further examine the role of LLM-based writing tools in the homogenization trend observed in Section~\ref{sec:nli_experiment}, we conducted a controlled rewriting experiment. We apply LLM-based rewriting to abstracts from all three eras, then measure the resulting change in NLI performance. This allows us to examine whether style normalization alone is sufficient to reduce L1 signals, independent of other factors such as domain shifts or changes in the author population.

\begin{table*}[t]
\centering
\small
\renewcommand{\arraystretch}{0.95}
\begin{tabular}{lllcccc}
\specialrule{1.5pt}{1pt}{2pt}
\textbf{Era} & \textbf{Classifier} & \textbf{Rewriter} & \textbf{Precision} &
\textbf{Recall} & \textbf{F1} & \textbf{$\Delta$F1} \\
\midrule
\multirow{6}{*}{pre-NN}
 & \multirow{3}{*}{Qwen3-14B}
   & ---            & 0.731 & 0.727 & 0.726 & ---      \\
 & & Gemma-3-12B-it & 0.515 & 0.380 & 0.372 & $-$0.354 \\
 & & GPT-4o         & 0.523 & 0.475 & 0.462 & $-$0.264 \\
\cmidrule{2-7}
 & \multirow{3}{*}{Gemma-3-12B-it}
   & ---            & 0.718 & 0.718 & 0.715 & ---      \\
 & & Qwen3-14B      & 0.608 & 0.574 & 0.569 & $-$0.146 \\
 & & GPT-4o         & 0.578 & 0.535 & 0.528 & $-$0.187 \\
\midrule
\multirow{6}{*}{pre-LLM}
 & \multirow{3}{*}{Qwen3-14B}
   & ---            & 0.671 & 0.640 & 0.637 & ---      \\
 & & Gemma-3-12B-it & 0.566 & 0.362 & 0.348 & $-$0.289 \\
 & & GPT-4o         & 0.561 & 0.446 & 0.436 & $-$0.201 \\
\cmidrule{2-7}
 & \multirow{3}{*}{Gemma-3-12B-it}
   & ---            & 0.643 & 0.628 & 0.614 & ---      \\
 & & Qwen3-14B      & 0.537 & 0.463 & 0.456 & $-$0.158 \\
 & & GPT-4o         & 0.524 & 0.417 & 0.413 & $-$0.201 \\
\midrule
\multirow{6}{*}{post-LLM}
 & \multirow{3}{*}{Qwen3-14B}
   & ---            & 0.721 & 0.633 & 0.620 & ---      \\
 & & Gemma-3-12B-it & 0.589 & 0.426 & 0.410 & $-$0.213 \\
 & & GPT-4o         & 0.572 & 0.428 & 0.396 & $-$0.227 \\
\cmidrule{2-7}
 & \multirow{3}{*}{Gemma-3-12B-it}
   & ---            & 0.669 & 0.615 & 0.598 & ---      \\
 & & Qwen3-14B      & 0.630 & 0.374 & 0.335 & $-$0.263 \\
 & & GPT-4o         & 0.611 & 0.400 & 0.400 & $-$0.198 \\
\specialrule{1.5pt}{1pt}{2pt}
\end{tabular}
\caption{Macro-averaged precision, recall, and F1-score before and after LLM
rewriting, averaged across three rewriting prompts. Rows where Rewriter is
\textbf{---} show the original fine-tuned scores before rewriting.
$\Delta$F1 is computed relative to the corresponding baseline.}
\label{tab:rewrite_macro}
\end{table*}

% % === Qwen3-14B classifier (main paper) ===
\begin{table*}[t]
\centering
\small
\setlength{\tabcolsep}{3.0pt}
\renewcommand{\arraystretch}{0.95}
\begin{tabular}{@{ }l|cccc c|cccc c|cccc c@{ }}
\specialrule{1.5pt}{1pt}{2pt}
& \multicolumn{5}{c|}{\textbf{pre-NN}}
& \multicolumn{5}{c|}{\textbf{pre-LLM}}
& \multicolumn{5}{c}{\textbf{post-LLM}} \\
& \multicolumn{2}{c}{P} & \multicolumn{2}{c}{R} & \multirow{2}{*}{$\Delta$F1}
& \multicolumn{2}{c}{P} & \multicolumn{2}{c}{R} & \multirow{2}{*}{$\Delta$F1}
& \multicolumn{2}{c}{P} & \multicolumn{2}{c}{R} & \multirow{2}{*}{$\Delta$F1} \\
\textbf{Lang.}
  & Bef. & Aft. & Bef. & Aft. &
  & Bef. & Aft. & Bef. & Aft. &
  & Bef. & Aft. & Bef. & Aft. & \\
\midrule
en-US & 0.603 & 0.336 & 0.700 & 0.767 & $-$0.181 & 0.569 & 0.296 & 0.500 & 0.620 & $-$0.171 & 0.471 & 0.376 & 0.800 & 0.727 & $-$0.096 \\
en-GB & 0.651 & 0.303 & 0.560 & 0.113 & $-$0.439 & 0.696 & 0.476 & 0.320 & 0.073 & $-$0.312 & 0.737 & 0.429 & 0.280 & 0.047 & $-$0.322 \\
fr    & 0.781 & 0.553 & 0.640 & 0.260 & $-$0.346 & 0.720 & 0.770 & 0.720 & 0.327 & $-$0.262 & 0.812 & 0.603 & 0.600 & 0.127 & $-$0.480 \\
de    & 0.673 & 0.500 & 0.700 & 0.487 & $-$0.193 & 0.630 & 0.466 & 0.580 & 0.427 & $-$0.158 & 0.625 & 0.541 & 0.600 & 0.573 & $-$0.055 \\
it    & 0.695 & 0.593 & 0.820 & 0.460 & $-$0.234 & 0.661 & 0.721 & 0.820 & 0.473 & $-$0.163 & 0.781 & 0.722 & 0.640 & 0.287 & $-$0.294 \\
zh    & 0.836 & 0.916 & 0.920 & 0.527 & $-$0.203 & 0.759 & 0.840 & 0.880 & 0.493 & $-$0.192 & 0.878 & 0.891 & 0.860 & 0.553 & $-$0.186 \\
ja    & 0.800 & 0.536 & 0.720 & 0.420 & $-$0.287 & 0.808 & 0.602 & 0.420 & 0.313 & $-$0.138 & 1.000 & 0.749 & 0.300 & 0.213 & $-$0.129 \\
ko    & 0.809 & 0.447 & 0.760 & 0.767 & $-$0.220 & 0.524 & 0.320 & 0.880 & 0.840 & $-$0.193 & 0.462 & 0.265 & 0.980 & 0.893 & $-$0.221 \\
\hline
Avg.  & 0.731 & 0.523 & 0.727 & 0.475 & $-$0.263 & 0.671 & 0.561 & 0.640 & 0.446 & $-$0.199 & 0.721 & 0.572 & 0.633 & 0.428 & $-$0.223 \\
\specialrule{1.5pt}{1pt}{2pt}
\end{tabular}
\caption{Per-language precision (P) and recall (R) before and after GPT-4o rewriting
across all three eras, evaluated by Qwen3-14B. $\Delta$F1 is computed from F1 scores
derived from the per-era averaged P and R. Aft.\ denotes values averaged across
three rewriting prompts. Results for Gemma-3-12B-it are in
Appendix~\ref{app:rewrite_results}.}
\label{tab:rewrite_pr_qwen}
\end{table*}

\subsection{Experimental Setup}
\label{sec:rewrite_setup}
The rewriting experiment targets abstracts from all three eras in our evaluation set. For each abstract, we apply a prompt-based rewriting process using three LLMs, Qwen3-14B, Gemma-3-12B-it, and GPT-4o \cite{openai2024gpt4technicalreport}. We apply a constraint that no abstract is classified by the same model used to rewrite it. Specifically, abstracts rewritten by Qwen3-14B are evaluated only by the fine-tuned Gemma-3-12B-it classifier, and vice versa. Abstracts rewritten by GPT-4o are evaluated by both fine-tuned classifiers.

We experiment with three rewriting prompts of increasing intervention strength, ranging from light proofreading to full fluency improvement. The full prompts are provided in Appendix~\ref{app:rewrite_prompt}. For each condition, we calculate the average precision and recall across the three prompts. We then recompute the F1 score from these averaged values to obtain a single result for each rewriter--classifier--era combination.

\subsection{Results}
\label{sec:rewrite_results}

Table~\ref{tab:rewrite_macro} presents the macro-averaged results before and after rewriting across all conditions. The results show a consistent drop in NLI performance after rewriting across all eras, classifiers, and rewriting models.

On pre-NN data, post-rewriting F1 ranges from 0.372 to 0.569 depending on the rewriter and classifier. This range is comparable to the NLI performance observed on actual post-LLM era text in Section~\ref{sec:nli_experiment}. The results suggest that LLM rewriting can reduce L1 signals in older abstracts to levels comparable to those seen in post-LLM academic writing that occurs naturally. On pre-LLM data, post-rewriting F1 ranges from 0.348 to 0.456. The smaller drop compared to pre-NN is expected as pre-LLM texts already carry weaker L1 signals and leave less room for further degradation.

On post-LLM data, NLI performance also drops after rewriting, with F1 decreasing by up to 0.263. This suggests that LLM rewriting can further reduce L1 signals even in texts that already reflect the homogenization trend. The effects also vary across rewriting models. Gemma-3-12B-it tends to cause larger drops than GPT-4o in several conditions. The bigger drop suggests that the degree of homogenization depends on model-specific tendencies rather than model size.

\paragraph{Per-language analysis.}
\label{sec:rewrite_language_analysis}

Table~\ref{tab:rewrite_pr_qwen} presents per-language precision and recall before and after GPT-4o rewriting that were evaluated by Qwen3-14B across all eras. The changes vary across language groups.

For British English (en-GB), recall drops to as low as 0.113 after rewriting in the pre-NN era. This indicates that the classifier rarely assigns the British English label to rewritten texts. At the same time, recall for American English (en-US) increases after rewriting across all eras. The increasing recall suggests that texts from other language groups are being reclassified as American English. This is consistent with the tendency of writing tools to lean toward American English conventions \cite{naturebias2025}.

For Chinese (zh), precision increases after rewriting while recall drops across all eras. This means the classifier becomes more selective but misses more Chinese-authored abstracts overall, implying that rewriting reduces many Chinese L1 features while the most distinctive ones stay detectable. This is consistent with Chinese showing the smallest overall drops in Table~\ref{tab:rewrite_macro} and with its resistance to homogenization observed in Section~\ref{sec:nli_experiment}.

\subsection{Discussion}
\label{sec:rewrite_discussion}
The rewriting experiment provides additional evidence that LLM-based writing tools can advance the homogenization trend observed in Section~\ref{sec:nli_experiment}. The drops after rewriting are larger than the differences between eras. This holds even for post-LLM texts that already carry weaker L1 signals. However, the experiment reflects controlled conditions. A single LLM applied uniformly to all abstracts does not reflect how researchers actually use these tools. In practice, not all authors use LLMs for writing, and those who do likely apply them selectively rather than rewriting entire abstracts. Therefore, the rewriting results represent an upper bound rather than an accurate representation of current practice. Nevertheless, as LLM-based writing assistance becomes increasingly widespread, the degree of homogenization may eventually reach the level observed in our experiment. Beyond direct tool use, broader familiarity with LLM-generated text may also gradually influence writing norms, though investigating this indirect effect is outside the scope of this work.

\section{Conclusion}
\label{sec:conclusion}

In this work, we investigated the impact of LLM-based writing assistance on the linguistic diversity of scientific writing. We constructed two NLI datasets for high-fluency academic writing and fine-tuned classifiers to detect L1 fingerprints across three periods. Our analysis reveals a consistent decline in NLI performance over time with F1-scores dropping by 10--12 percentage points from pre-NN to post-LLM. Notably, the most significant drop occurs between the pre-NN and pre-LLM eras, coinciding with the rise of neural machine translation around 2016. The shift into the post-LLM era is smaller and not statistically significant. These results suggest that writing homogenization has been a gradual process that precedes LLMs.

What LLMs appear to have done is amplify this trend. Our rewriting experiment shows that LLM-based rewriting reduces NLI performance across all eras, including post-LLM texts that already carry weaker L1 signals, to a degree that exceeds the natural era progression. Overall, LLMs contribute to homogenization, but the trend itself appears to have started before them, with the largest shift coinciding with the rise of NMT. Chinese maintains stable or improving detectability across eras, possibly related to the distinct Chinese AI ecosystem, while French shows mixed trends possibly reflecting the availability of non-English-dominant models in the European NLP community. These differences suggest that homogenization is shaped not only by linguistic structure but also by the broader technological environment surrounding each speaker community.

Future work could extend this analysis by broadening the language and domain coverage. Additionally, investigating whether similar homogenization trends emerge in non-academic writing would help determine whether this occurrence is a broader effect of LLM adoption. It would also clarify whether the trend is specific to academic publishing.

\section*{Limitations}
Our L1 labels are high-probability estimates rather than ground truth. Filtering by author name and affiliation introduces non-trivial sampling bias that excludes non-Anglo names from English-native pools. The methodology also cannot account for complex linguistic backgrounds such as heritage speakers or multilingual researchers. While we acknowledge these limitations, we argue that the bias is unlikely to affect our main conclusion. More robust labeling strategies that would improve data quality, such as tracking author background through CVs or institutional profiles, are left for future work.

We define the ``post-LLM" era (2023-2025) as the era of heavy AI influence. However, we cannot verify whether every author whose data was included in this set actually utilized LLMs. \citet{liang2024mappingincreasingusellms} document a measurable surge in LLM-marker vocabulary in scientific literature beginning in late 2023, consistent with our era boundary. However, actual usage patterns likely vary across authors. The homogenization we observe may reflect direct LLM use or a continuation of trends that started from NMT tools. Our statistical significance results further suggest that the most significant shift occurred between the pre-NN and pre-LLM eras. This indicates the trend precedes widespread LLM adoption.

Our dataset is constructed exclusively from computer science and NLP publications (arXiv/ACL). As this field is among the earliest adopters of AI tools, the homogenization trends we observe may be more pronounced here than in other disciplines (e.g., humanities or social sciences) where AI adoption might be slower or deployed differently. Additionally, our analysis is restricted to paper abstracts that tend to be the most carefully edited part of a paper. This may cause the abstracts to exhibit less, or perhaps more L1 variation than body text, depending on actual writing tool use by the authors.

Our era-based comparison may also be affected by confounding factors that are not directly related to writing assistance tools. These include broader multilingual media exposure over time, the substantial growth in the volume of scientific writing published per year, and shifts in topic distribution across eras. Any of these factors could independently influence surface-level writing style and contribute to the NLI performance decline we observe, alongside the homogenization effect we attribute to writing assistance tools. We do not attempt to fully discuss these factors in this work. We treat them as an important direction for future investigation.

The rewriting experiment covers three prompt-based rewriting scenarios ranging from light proofreading to full fluency improvement. However, these do not cover all real-world use cases of LLM writing assistance that may involve a more iterative and selective editing. Finally, our evaluation set consists of 50 papers per language per era, which is relatively small for drawing strong statistical conclusions. For underrepresented combinations, we address this by duplicating a subset of collected papers. As a result, this may introduce a small bias in the results for those specific cells.

\section*{Acknowledgments}
This work was partly supported by JSPS KAKENHI Grant Numbers 24H00727.

\bibliography{custom}

\appendix

\section{Linguistic Features Analysis}
\label{app:features_analysis}

\subsection{Chosen Features}
\label{app:chosen_features}
Feature selection was made based on two criteria. First, the feature must be theoretically supported by literature related to L1 transfer patterns in English. It must also be extractable from short texts using already available NLP tools without requiring manual annotation.

We drew primarily on \citet{swansmith2001learners}, a comprehensive reference on L1 interference patterns in English across a wide range of native language, not limited to writing but also spoken language and vocabulary. For each of the eight languages included in our study, we identified grammatical and syntactic patterns known to transfer from L1 into English. Some examples include article omission in East Asian language writers and passive voice preferences in European language writers. Next, we selected features that can be reliably extracted using spaCy \cite{Honnibal2020Spacy}, with the \texttt{en\_core\_web\_lg} model (v3.8.3). All features are computed as normalized ratios to control for variation in abstract length in our dataset.

Initially, we identified 20 candidate features that are listed in Table~\ref{tab:app_features}. However, when visualized as heatmaps, the per-language trends were difficult to interpret due to the high dimension. Therefore, we selected a reduced set of 9 features for the main trend analysis.

\begin{table*}[t]
\centering
\small
\renewcommand{\arraystretch}{1.0}
\begin{tabular}{clcp{4.8cm} p{4.0cm}}
\specialrule{1.5pt}{1pt}{2pt}
\textbf{\#} & \textbf{Feature} & \textbf{Used} & \textbf{Definition} & \textbf{Motivation} \\
\hline
1  & Passive voice frequency      & \Checkmark & Passive constructions / sentences & Universal across L1 groups \\ \hline 
2  & Non-finite form preference   &            & Gerund : infinitive : nominalization ratio per sentence & Distinct patterns across all non-English L1s \\\hline 
3  & Relative clause frequency    & \Checkmark & Relative clauses / sentences & Distinct patterns across all L1s \\\hline 
4  & Article usage frequency      & \Checkmark & Articles (a, an, the) / total words & Separates European from East Asian L1s \\\hline 
5  & Modal verb frequency         & \Checkmark & Modal verbs / total verbs & Distinct patterns across all L1s \\\hline 
6  & Adverbial placement          &            & Pre-verbal vs.\ post-verbal adverbial ratio & Distinctive for German, Chinese \\\hline 
7  & Noun compound frequency      &            & Noun-noun compounds / total nouns & Distinctive for German, Japanese \\\hline 
8  & Subordinate clause ratio     & \Checkmark & Subordinate clauses / sentences & Distinctive across language families \\\hline 
9  & Sentence length              & \Checkmark & Mean words per sentence & Universal baseline feature \\\hline 
10 & Topic-fronting frequency     &            & Non-subject-initial sentences / total sentences & Distinctive for East Asian languages \\\hline 
11 & Number marking frequency     &            & Plural nouns / total nouns + 3rd-person \textit{-s} presence & Distinctive for East Asian languages \\\hline 
12 & Hedging expression frequency & \Checkmark & Hedging words / total words (\textit{may, might, possibly}, etc.) & Distinctive for Japanese \\\hline 
13 & Conjunction patterns         &            & Conjunctions / sentences + duplicate conjunction detection & Distinctive for Chinese, Korean \\\hline 
14 & Phrasal verb frequency       &            & Phrasal verbs / total verbs & Distinctive for East Asian languages \\\hline 
15 & Perfect aspect frequency     &            & Perfect constructions / total verbs & Distinctive for Korean \\\hline 
16 & Progressive aspect frequency &            & Progressive constructions / total verbs & Distinctive across multiple L1s \\\hline 
17 & Pronoun frequency            & \Checkmark & Pronouns / total words & Distinctive for East Asian languages \\\hline 
18 & Preposition variety          &            & Unique prepositions / total prepositions (TTR) & Distinctive for Chinese, French \\\hline 
19 & Clause length                &            & Mean words per clause & Complements sentence length \\\hline 
20 & Lexical density              &            & Content words (N, V, Adj, Adv) / total words & Captures overall writing style \\
\specialrule{1.5pt}{1pt}{2pt}
\end{tabular}
\caption{Linguistic features used for analysis. Checkmarks (\Checkmark) indicate
the 9 features selected for the trend analysis in Figure~\ref{fig:feature_heatmap_comparison}}
\label{tab:app_features}
\end{table*}

\subsection{Feature Heatmap}
\label{app:feature_heatmap}
\begin{figure*}[h]
    \centering
    \includegraphics[width=0.8\linewidth]{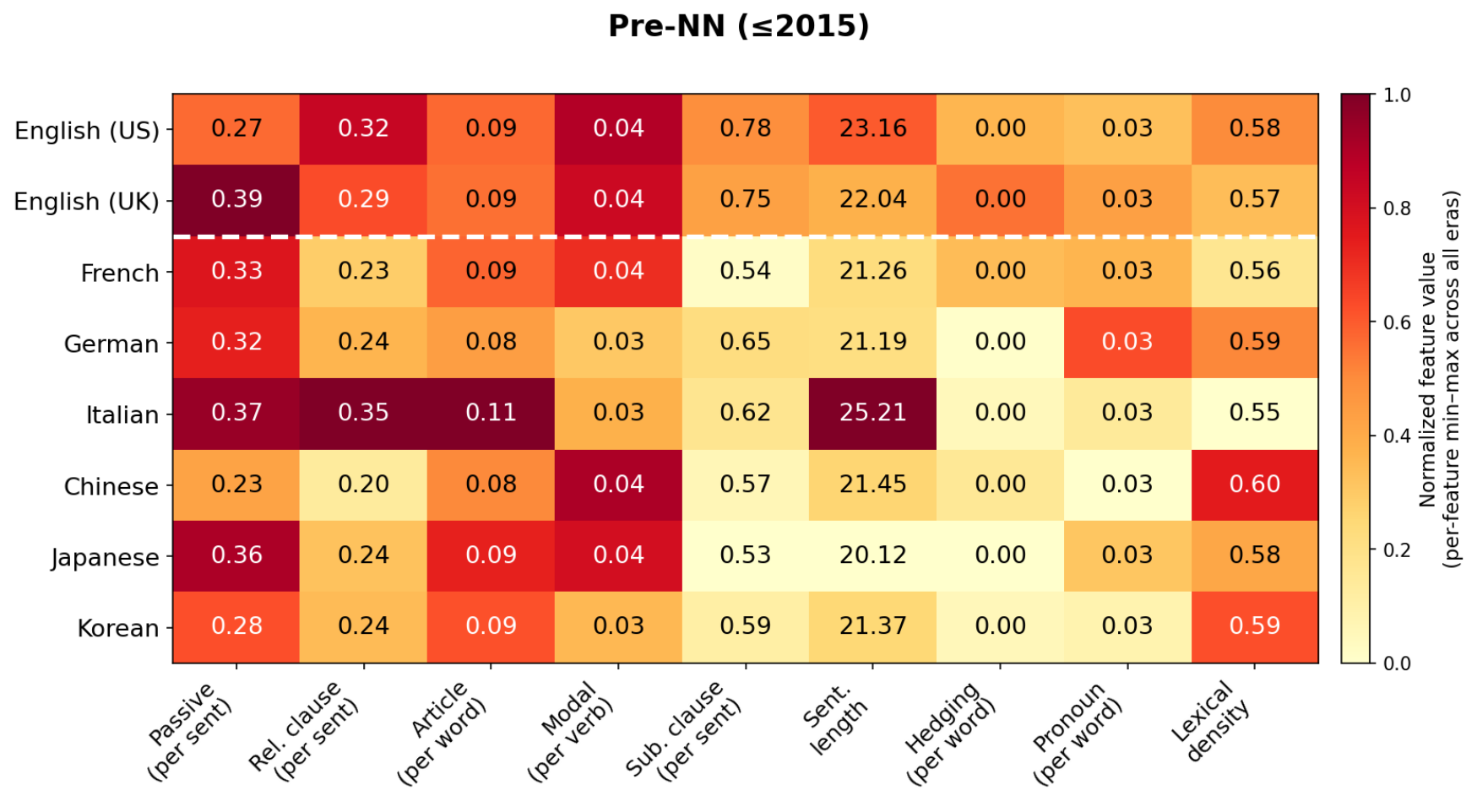}
    \vspace{0.5em}
    \includegraphics[width=0.8\linewidth]{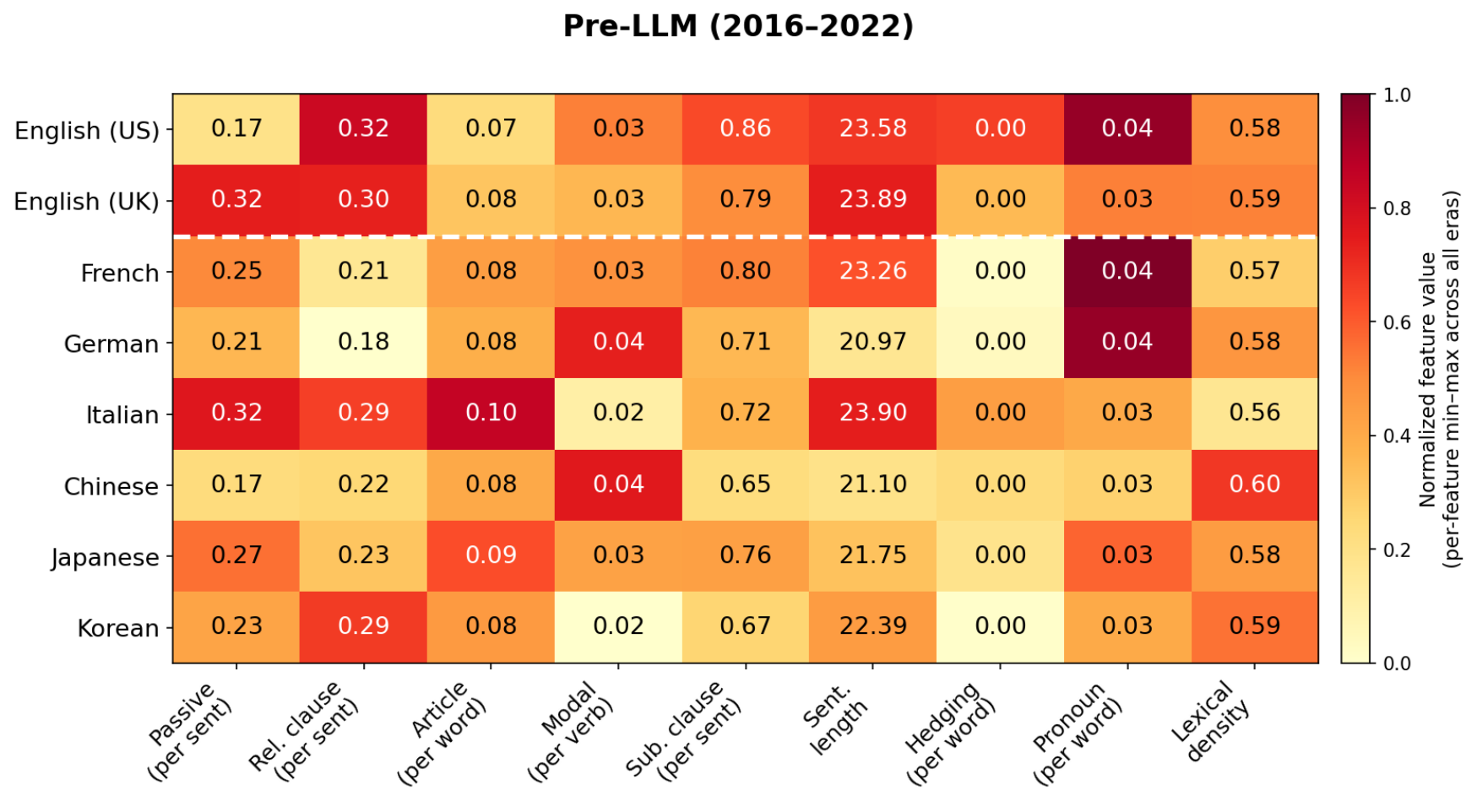}
    \vspace{0.5em}
    \includegraphics[width=0.8\linewidth]{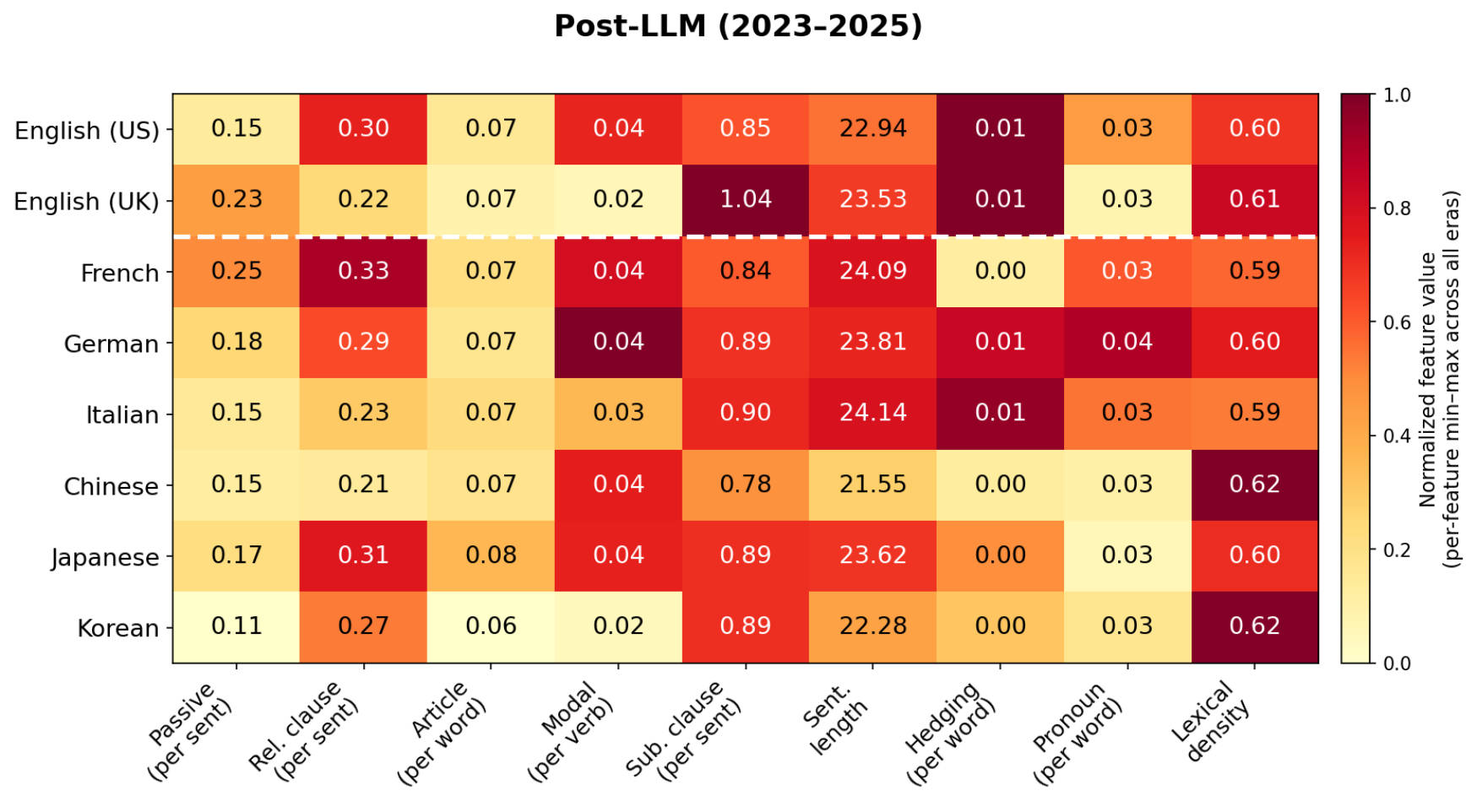}
    \caption{Comparison of feature analysis heatmaps across Pre-NN, Pre-LLM, and Post-LLM datasets.}
    \label{fig:feature_heatmap_comparison}
\end{figure*}

Figure~\ref{fig:feature_heatmap_comparison} presents the mean values of the 9 selected features per language group across the three eras, with colors normalized per feature across all eras to highlight relative differences rather than absolute values. In the pre-NN era, different feature profiles can clearly be seen in non-native language groups. This reflects the L1-specific transfer tendencies documented in \citet{swansmith2001learners}. East Asian groups (Japanese, Chinese, Korean) show lower article usage frequency that is consistent with the grammar of respective languages. On the other hand, European groups (French, German, Italian) have higher subordinate clause ratios. This aligns with the more complex clause structures in their native grammar. As we move into pre-LLM and post-LLM, the differences begin to narrow and converge toward American English norms. Interestingly, British English also exhibits a stylistic shift despite being a native English variety.

\section{Few-shot Experiment}
\subsection{Prompt}
\label{app:appendix_fewshot_prompt}
To ensure the model focuses on stylistic interference rather than content, we employed the following system prompt for few-shot inference tasks.\newpage

\vspace{1em}

\begin{quote}
\small
\textbf{System Prompt:} You are an expert computational linguist specializing in Native Language Identification (NLI).
Task: Identify the author’s native language by detecting L1-interference patterns in English writing.

\vspace{0.3em}
\noindent \textbf{Important decision rule:}
\begin{itemize}
    \setlength\itemsep{0em}
    \item Do NOT choose an English native label unless there is strong positive evidence (e.g., consistent British/American spelling, idiomatic phrasing, no L1 interference).
    \item High fluency alone is NOT evidence of native English.
    \item If any systematic L1-interference is present, prefer a non-English label.
\end{itemize}

\noindent \textbf{Valid labels:} \texttt{english\_american}, \texttt{english\_british}, \texttt{french}, \texttt{german}, \texttt{italian}, \texttt{chinese}, \texttt{japanese}, \texttt{korean}

\vspace{0.3em}
\noindent \textbf{Constraint:} No explanations. Only select from the given labels. Other languages are NOT possible. Do not try answering with any other language because it is guaranteed to be FALSE.

\vspace{0.5em}
\noindent \textbf{User Message:} Classify the native language: \{Title\} \{Abstract\}
\end{quote}

\subsection{Full Results}
\label{sec:app_fewshot_results}

\begin{figure}[t]
    \centering
    \includegraphics[width=\columnwidth]{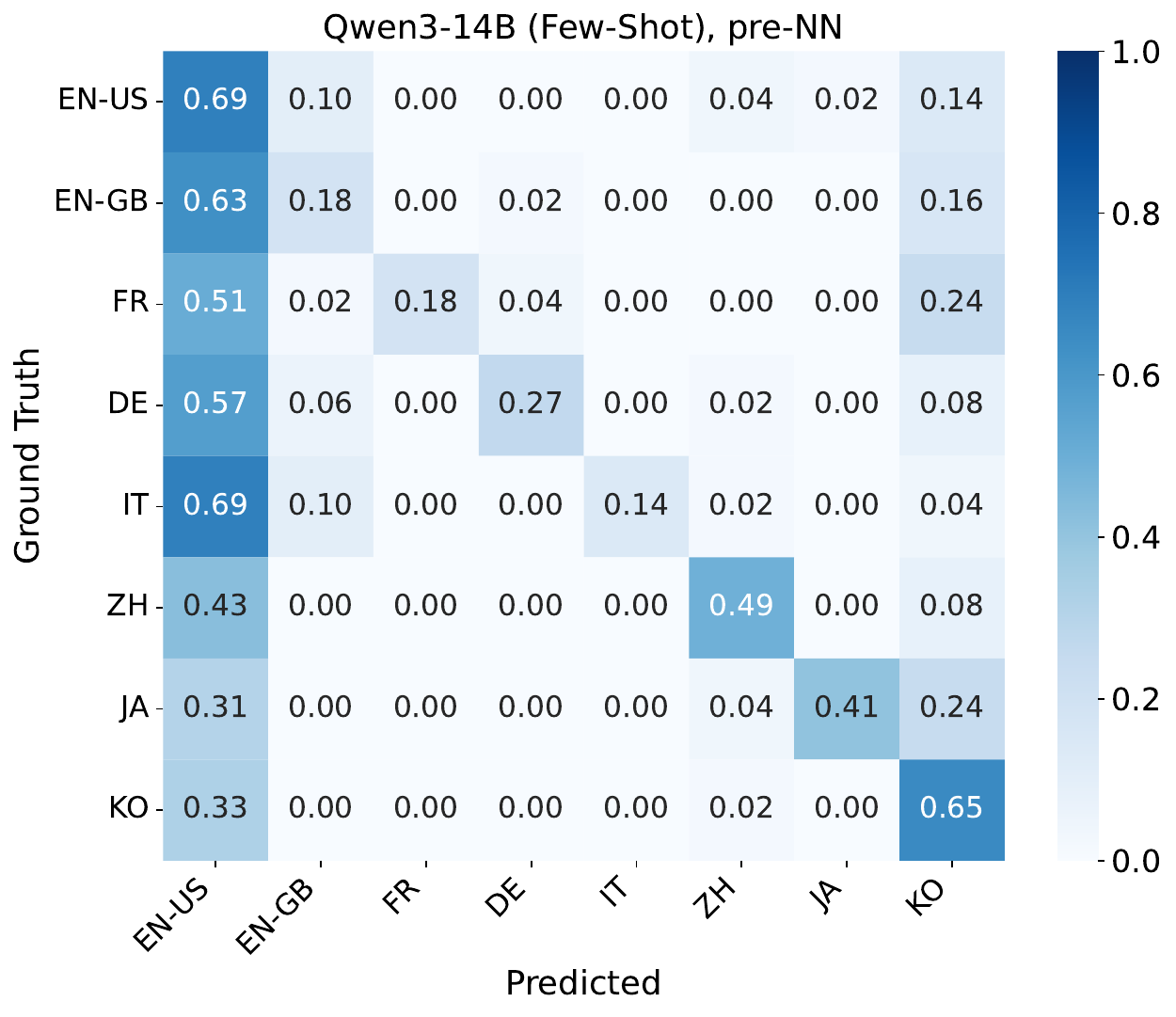}
    \caption{Confusion matrix for Qwen3-14B (few-shot, pre-NN).}
    \label{fig:cm_fewshot_qwen_prenn}
\end{figure}

\begin{figure}[t]
    \centering
    \includegraphics[width=\columnwidth]{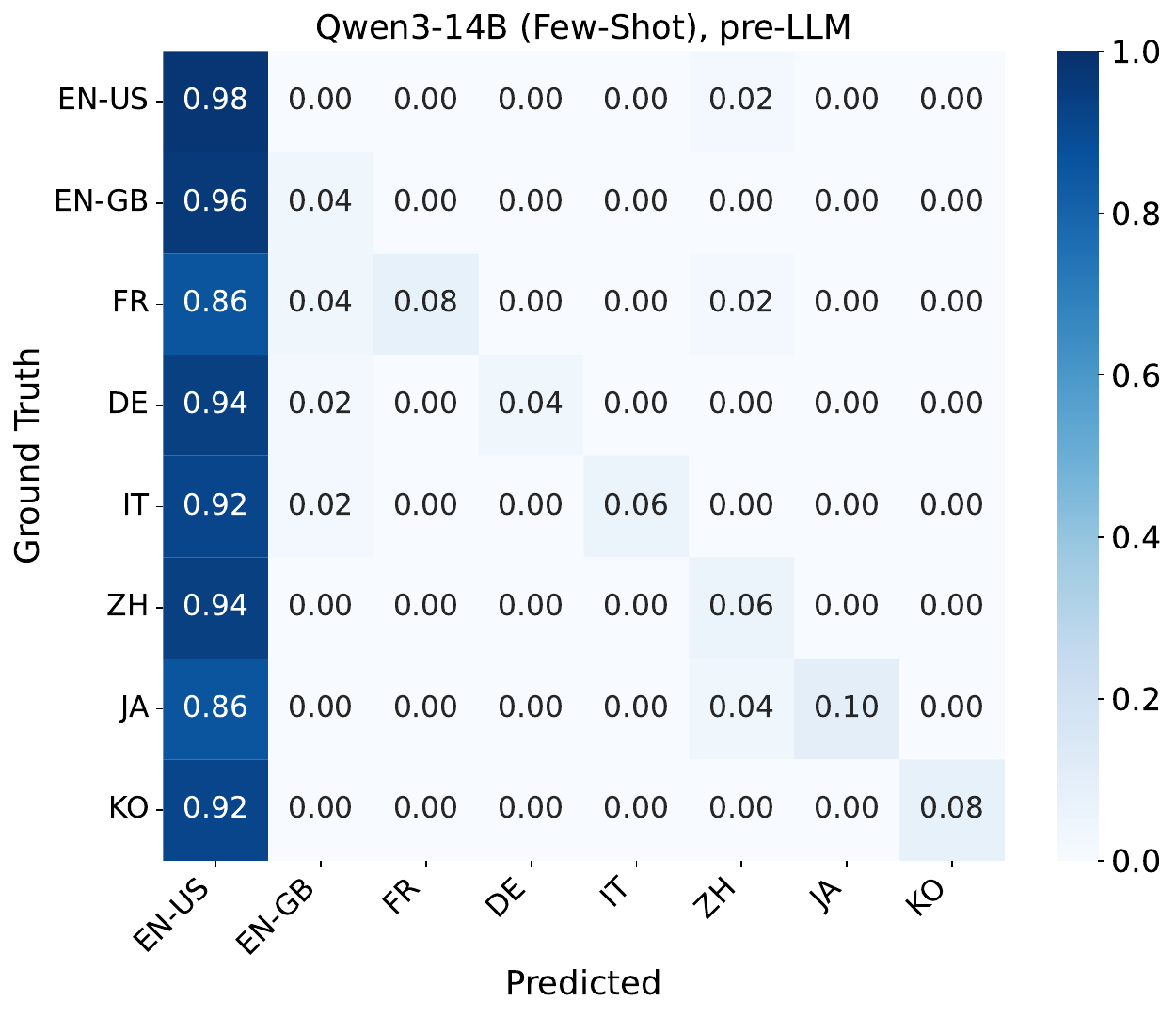}
    \caption{Confusion matrix for Qwen3-14B (few-shot, pre-LLM).}
    \label{fig:cm_fewshot_qwen_prellm}
\end{figure}

\begin{figure}[t]
    \centering
    \includegraphics[width=\columnwidth]{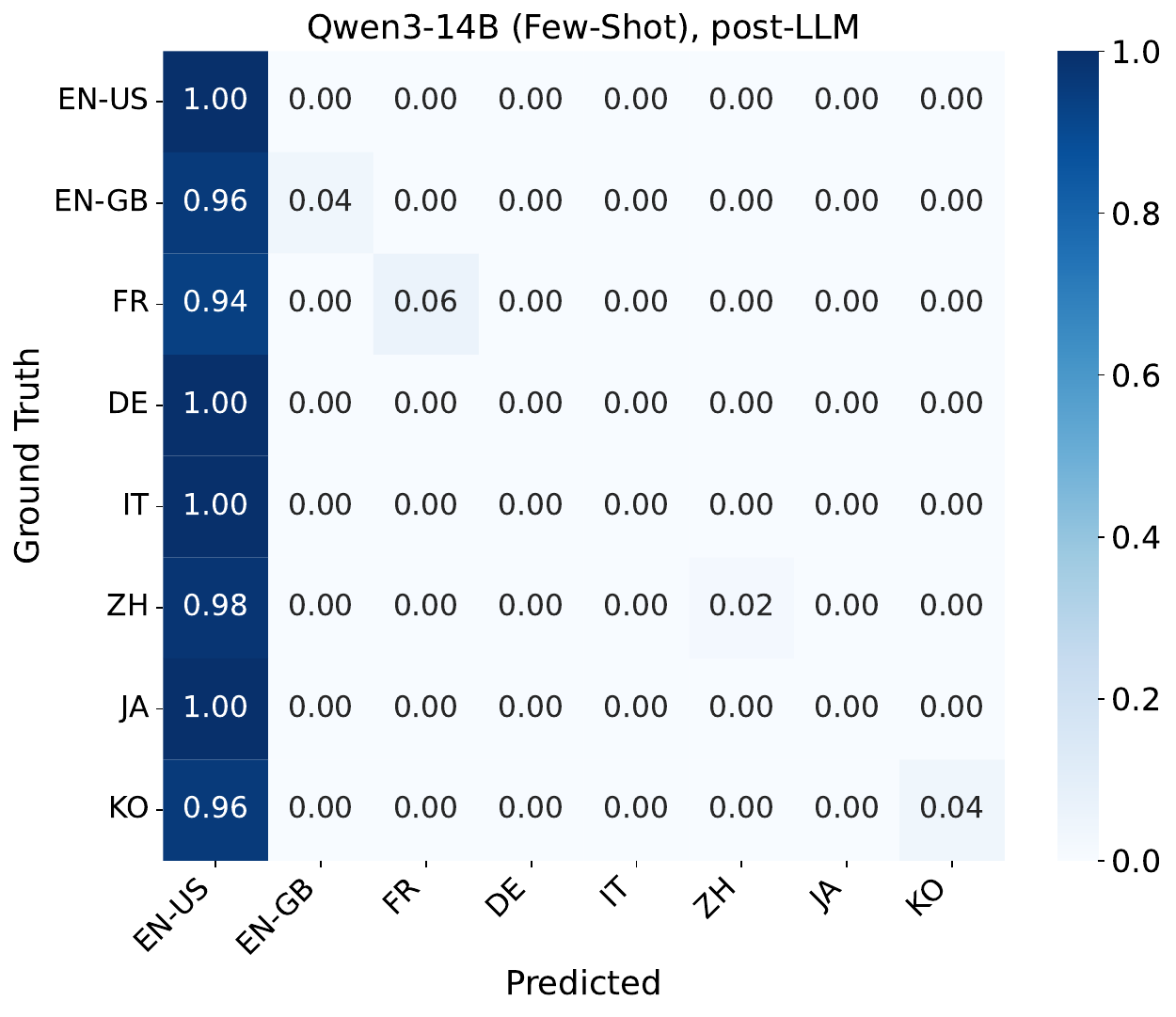}
    \caption{Confusion matrix for Qwen3-14B (few-shot, post-LLM).}
    \label{fig:cm_fewshot_qwen_postllm}
\end{figure}

\begin{figure}[t]
    \centering
    \includegraphics[width=\columnwidth]{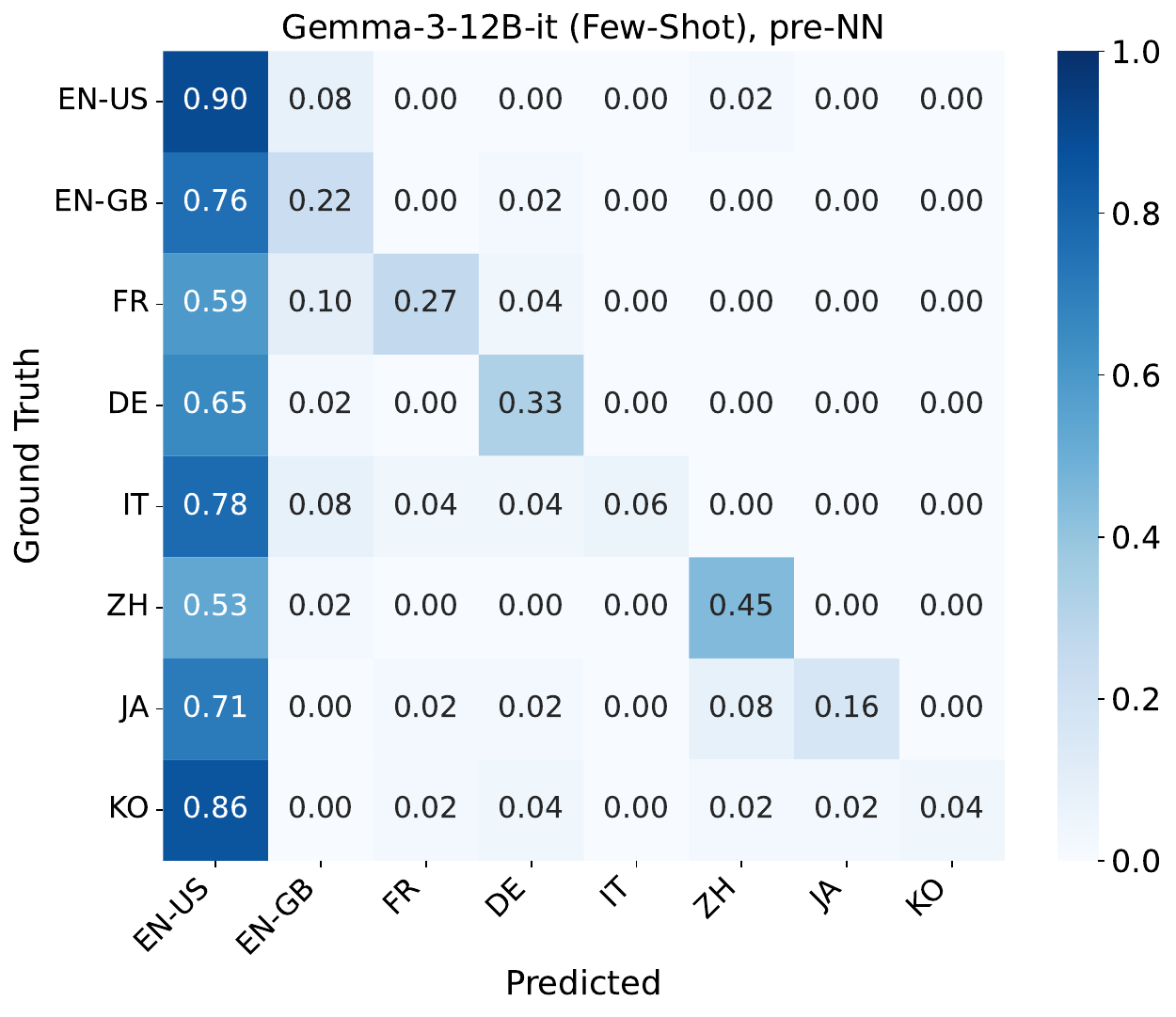}
    \caption{Confusion matrix for Gemma-3-12B-it (few-shot, pre-NN).}
    \label{fig:cm_fewshot_gemma_prenn}
\end{figure}

\begin{figure}[t]
    \centering
    \includegraphics[width=\columnwidth]{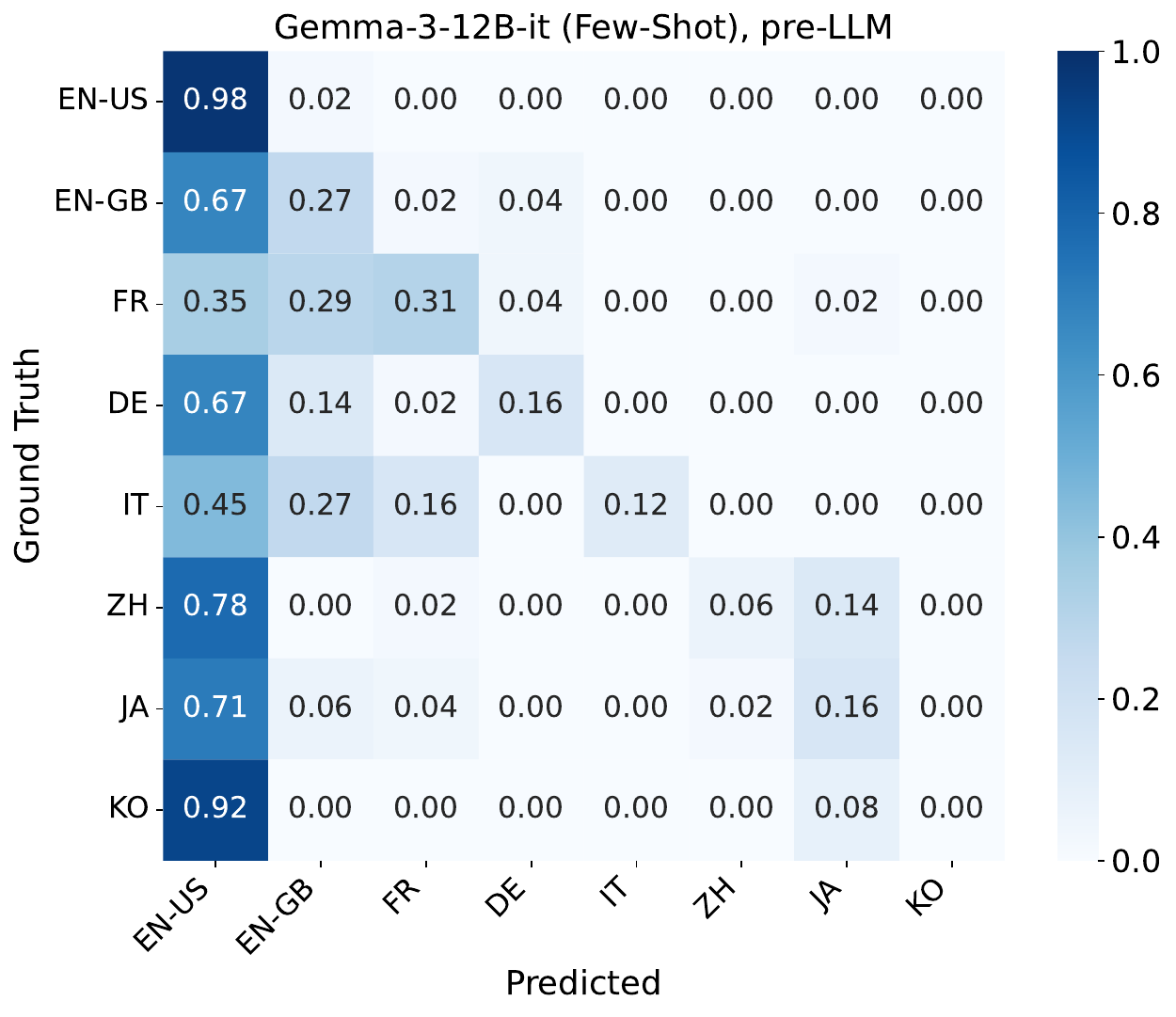}
    \caption{Confusion matrix for Gemma-3-12B-it (few-shot, pre-LLM).}
    \label{fig:cm_fewshot_gemma_prellm}
\end{figure}

\begin{figure}[t]
    \centering
    \includegraphics[width=\columnwidth]{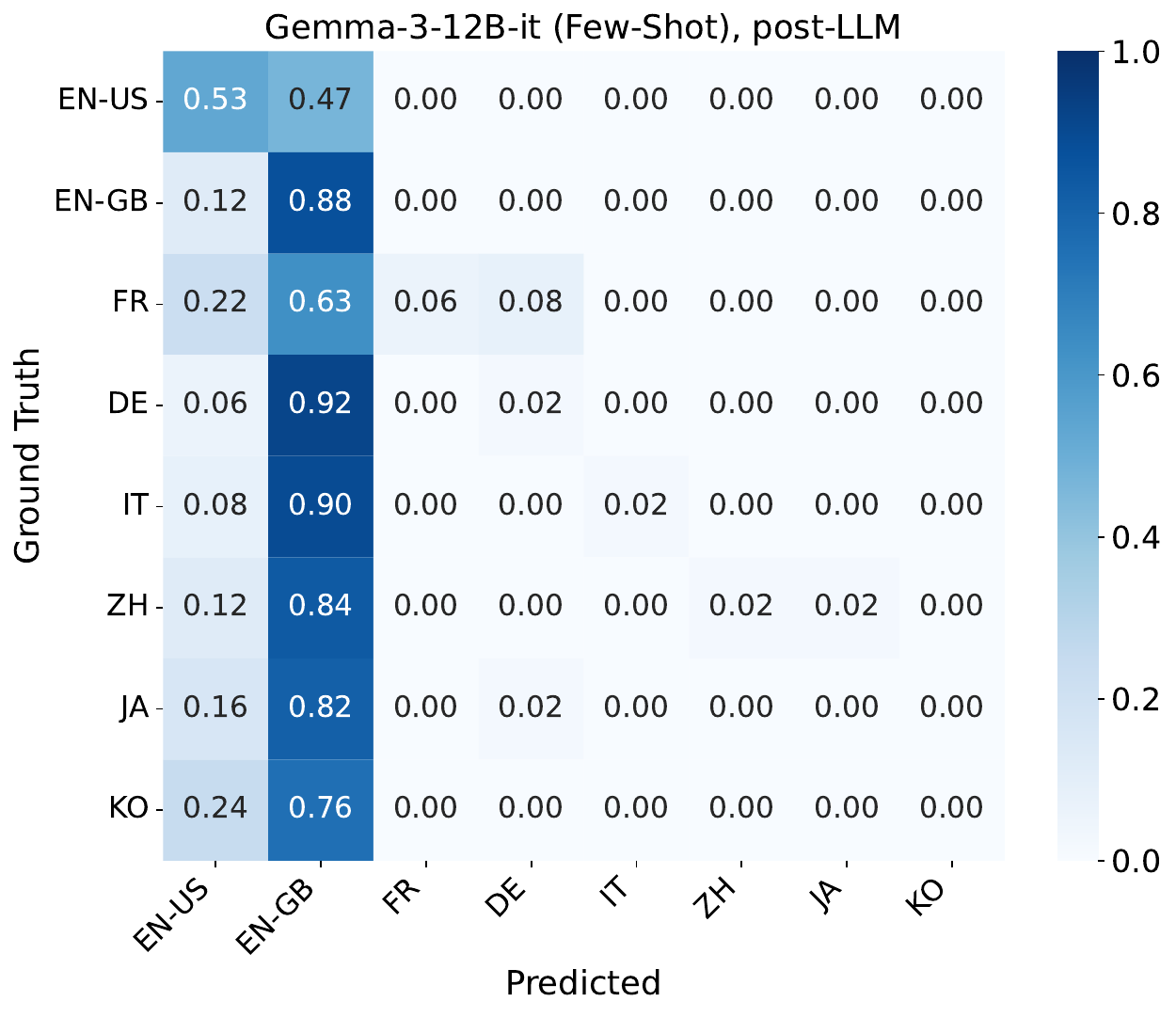}
    \caption{Confusion matrix for Gemma-3-12B-it (few-shot, post-LLM).}
    \label{fig:cm_fewshot_gemma_postllm}
\end{figure}

Figures \ref{fig:cm_fewshot_qwen_prenn}--\ref{fig:cm_fewshot_gemma_postllm} present the confusion matrices for the Qwen3 and Gemma 3 models in the few-shot setting. 

Unlike the fine-tuned models, the base models exhibit significant class imbalance in their predictions. In the pre-LLM and post-LLM eras, the models frequently default to predicting \texttt{english\_american} or \texttt{english\_british} regardless of the actual input. This results in high diagonal values for English classes but near-zero recall for other native languages. 

% --- Qwen Few-Shot: pre-NN ---

% --- Qwen Few-Shot: pre-LLM ---

% --- Qwen Few-Shot: post-LLM ---

% --- Gemma Few-Shot: pre-NN ---

% --- Gemma Few-Shot: pre-LLM ---

% --- Gemma Few-Shot: post-LLM ---

\section{Fine-Tuning}
\subsection{Hyperparameters for Fine-Tuning}
\label{app:appendix_hyperpara}
The specific hyperparameters we used to fine-tune each of the models are shown in Table \ref{tab:appendix_hyperpara}.
\begin{table}[h]
\centering
\small
% This command forces the table to fit exactly within the column width
\resizebox{\columnwidth}{!}{%
\begin{tabular}{lcc}
\specialrule{1.5pt}{1pt}{2pt}
\textbf{Hyperparameter} & \textbf{Qwen3-14B} & \textbf{Gemma-3-12B-it} \\ \hline
Epochs & 2 & 3 \\
Batch Size & 8 & 16 \\
Gradient Accumulation & 4 & 2 \\
Learning Rate & $1.0 \times 10^{-3}$ & $2.0 \times 10^{-4}$ \\
Lora Rank ($r$) & 16 & 16 \\
Lora Alpha ($\alpha$) & 64 & 32 \\
Lora Dropout & 0.0001 & 0.1 \\
Weight Decay & 0 & 0.01 \\
\specialrule{1.5pt}{1pt}{2pt}
\end{tabular}%
}
\caption{Hyperparameters for fine-tuning using QLoRA.}
\label{tab:appendix_hyperpara}
\end{table}

\subsection{Prompt}
\label{app:appendix_prompt_tuning}
We employed the following system prompt for all fine-tuning.
\vspace{1em}

\begin{quote}
\small
\textbf{System Prompt:} You are an expert linguist specializing in native-language identification from English writing. Your task is to identify the author's native language based purely on writing style, grammar, syntax, word choice, collocations, and subtle L1 interference patterns. Your output must be exactly one label, lowercase, from this list: \texttt{english\_american}, \texttt{english\_british}, \texttt{french}, \texttt{german}, \texttt{italian}, \texttt{chinese}, \texttt{japanese}, \texttt{korean}. Do not output any other languages except the provided ones. Do not output any additional explanations, thinking, or arguments, aside from the language name in lowercase. Example: french

\vspace{0.5em}
\noindent \textbf{User Message:} Analyze the following text and determine the author's native language. 
Text: \{Title\} \{Abstract\}
Native Language:
\end{quote}

\subsection{Full Results}
\label{app:ft_full_result}
\begin{figure}[t]
    \centering
    \includegraphics[width=\columnwidth]{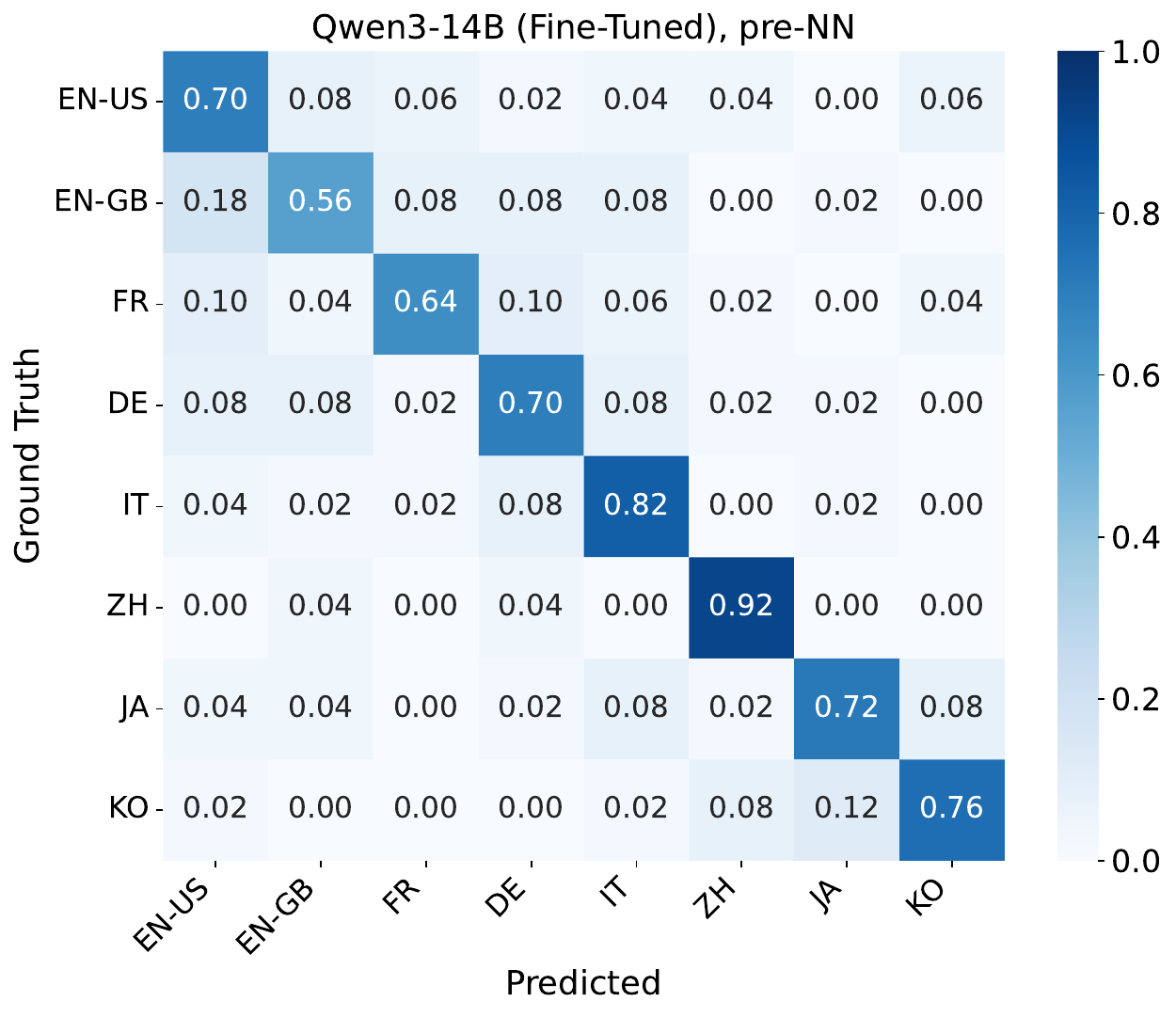}
    \caption{Confusion matrix for Qwen3-14B (fine-tuned, pre-NN).}
    \label{fig:cm_full_qwen_prenn}
\end{figure}

\begin{figure}[t]
    \centering
    \includegraphics[width=\columnwidth]{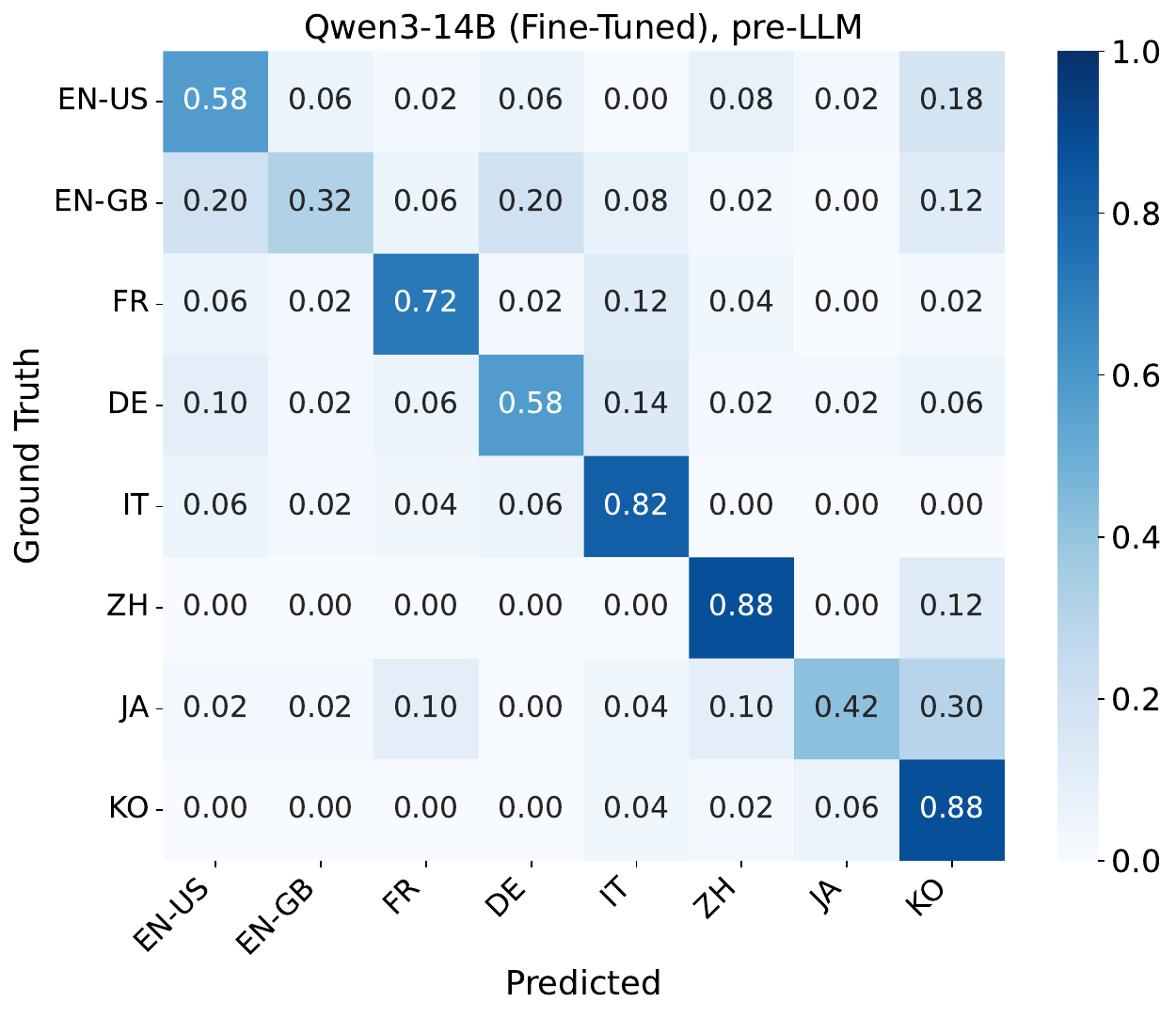}
    \caption{Confusion matrix for Qwen3-14B (fine-tuned, pre-LLM).}
    \label{fig:cm_full_qwen_prellm}
\end{figure}

\begin{figure}[t]
    \centering
    \includegraphics[width=\columnwidth]{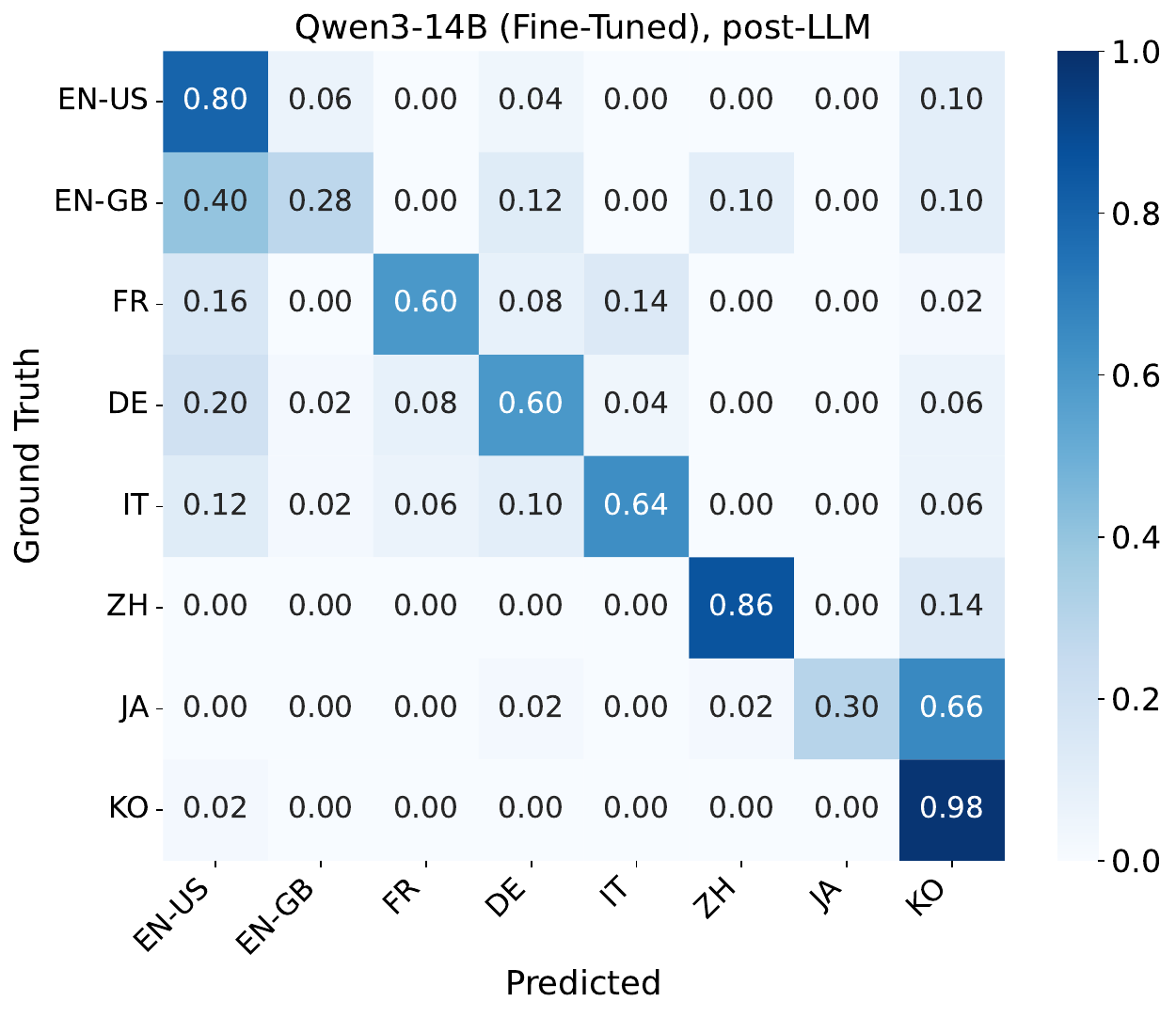}
    \caption{Confusion matrix for Qwen3-14B (fine-tuned, post-LLM).}
    \label{fig:cm_full_qwen_postllm}
\end{figure}

Table~\ref{tab:appendix_qwen_detailed_res} presents the full per-class precision, recall, and F1-score for Qwen3-14B across all three eras. A consistent pattern is the sharp decline in recall for Japanese and Korean in the post-LLM era. This suggests that the model becomes overly conservative by only predicting Japanese when very certain. Chinese maintains consistently high scores across all eras. Figures~\ref{fig:cm_full_qwen_prenn}--\ref{fig:cm_full_qwen_postllm} show the corresponding confusion matrices.

\begin{table*}[h]
\centering
\small
\setlength{\tabcolsep}{4pt}

\begin{tabular}{l|ccc|ccc|ccc}
\specialrule{1.5pt}{1pt}{2pt}
\textbf{Language} & \multicolumn{3}{c|}{\textbf{Precision}} & \multicolumn{3}{c|}{\textbf{Recall}} & \multicolumn{3}{c}{\textbf{F1-score}} \\
 & pre-NN & pre-LLM & post-LLM & pre-NN & pre-LLM & post-LLM & pre-NN & pre-LLM & post-LLM \\ \hline
en-US & 0.603& 0.569 & 0.471 & 0.700 & 0.500 & 0.800 & \textbf{0.648} & 0.574 & 0.593 \\
en-GB & 0.651 & 0.696 & 0.737 & 0.560 & 0.320 & 0.280 & \textbf{0.602} & 0.438 & 0.406 \\
fr & 0.781 & 0.720 & 0.812 & 0.640 & 0.720 & 0.600 & 0.703 & \textbf{ 0.720 }& 0.690 \\
de & 0.673 & 0.630 & 0.625 & 0.700 & 0.580 & 0.600 & \textbf{0.686} & 0.604 & 0.612 \\
it & 0.695 & 0.661 & 0.781 & 0.820 & 0.820 & 0.640 & \textbf{0.752} & 0.732 & 0.703 \\
zh & 0.836 & 0.759 & 0.878 & 0.920 & 0.880 & 0.860 & \textbf{0.876} & 0.815 & 0.869 \\
ja & 0.800 & 0.808 & 1.000 & 0.720 & 0.420 & 0.300 & \textbf{0.758} & 0.553 & 0.462 \\
ko & 0.809 & 0.524 & 0.462 & 0.760 & 0.880 & 0.980 & \textbf{0.784} & 0.657 & 0.628 \\ 
\specialrule{1.5pt}{1pt}{2pt}
\end{tabular}

\caption{Detailed per-class performance metrics for Qwen3-14B (fine-tuned).}
\label{tab:appendix_qwen_detailed_res}
\end{table*}

% --- Qwen Full: pre-NN ---

% --- Qwen Full: pre-LLM ---

% --- Qwen Full: post-LLM ---

\begin{figure}[t]
    \centering
    \includegraphics[width=\columnwidth]{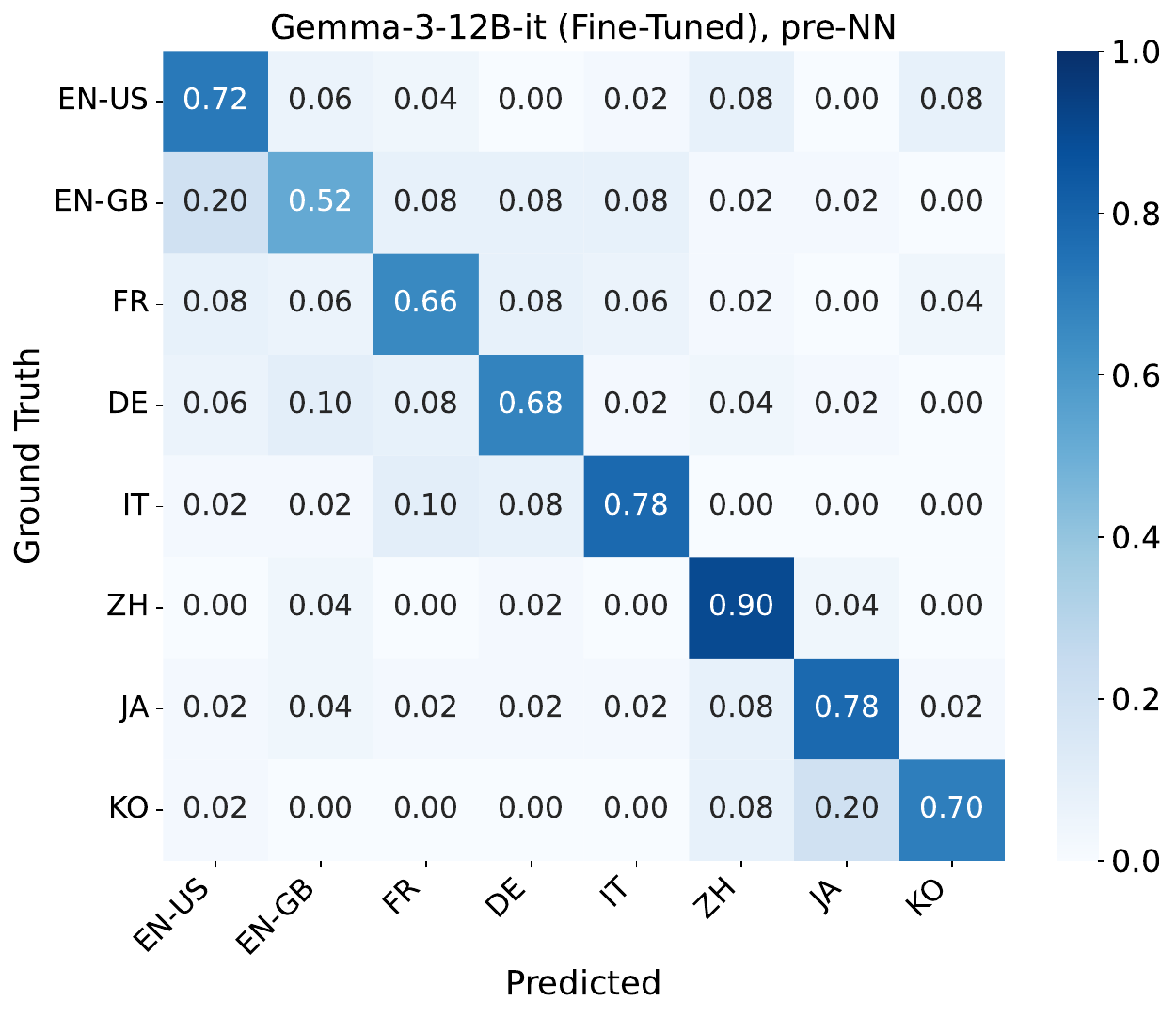}
    \caption{Confusion matrix for Gemma-3-12B-it (fine-tuned, pre-NN).}
    \label{fig:cm_full_gemma_prenn}
\end{figure}

\begin{figure}[t]
    \centering
    \includegraphics[width=\columnwidth]{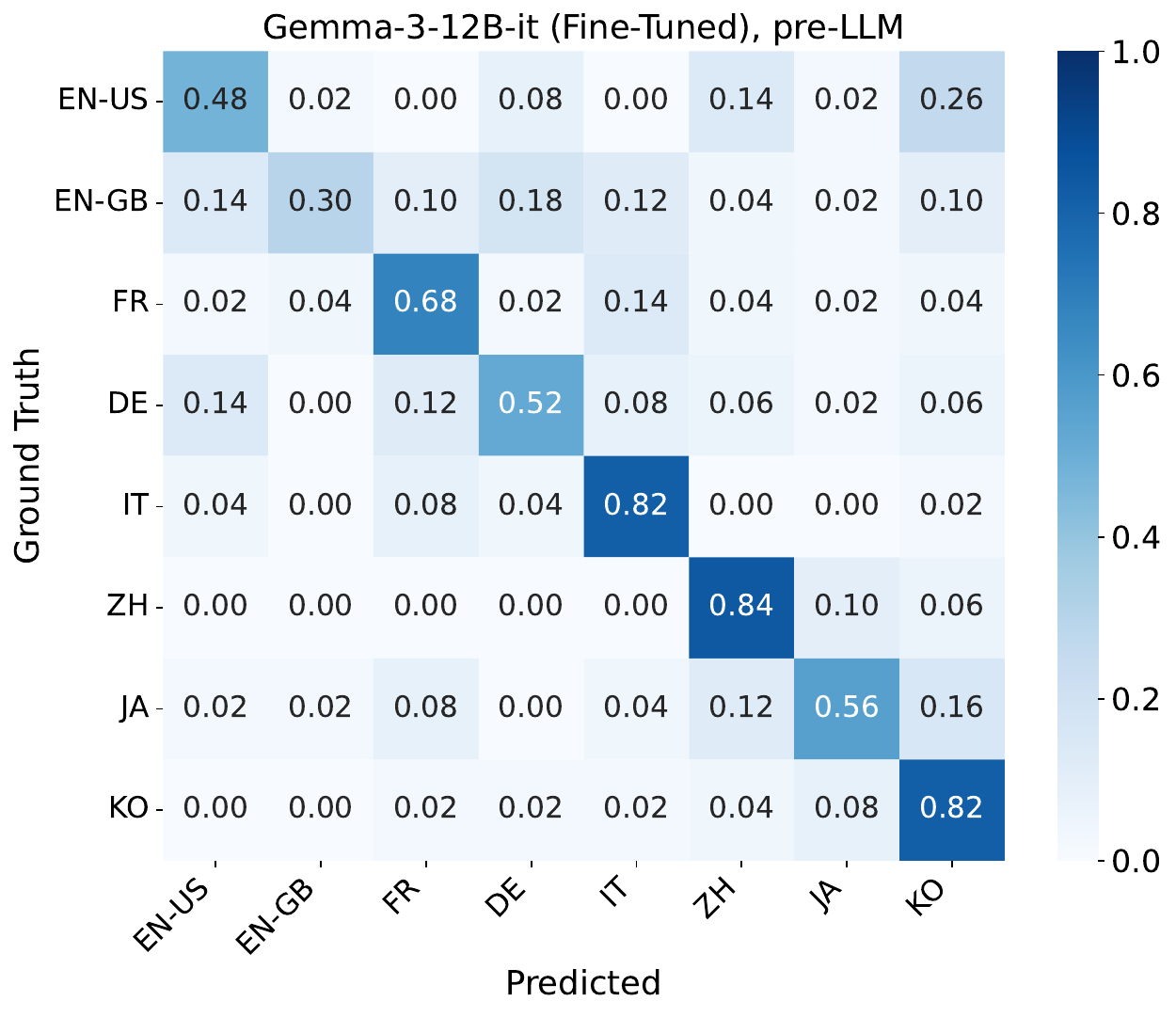}
    \caption{Confusion matrix for Gemma-3-12B-it (fine-tuned, pre-LLM).}
    \label{fig:cm_full_gemma_prellm}
\end{figure}

\begin{figure}[t]
    \centering
    \includegraphics[width=\columnwidth]{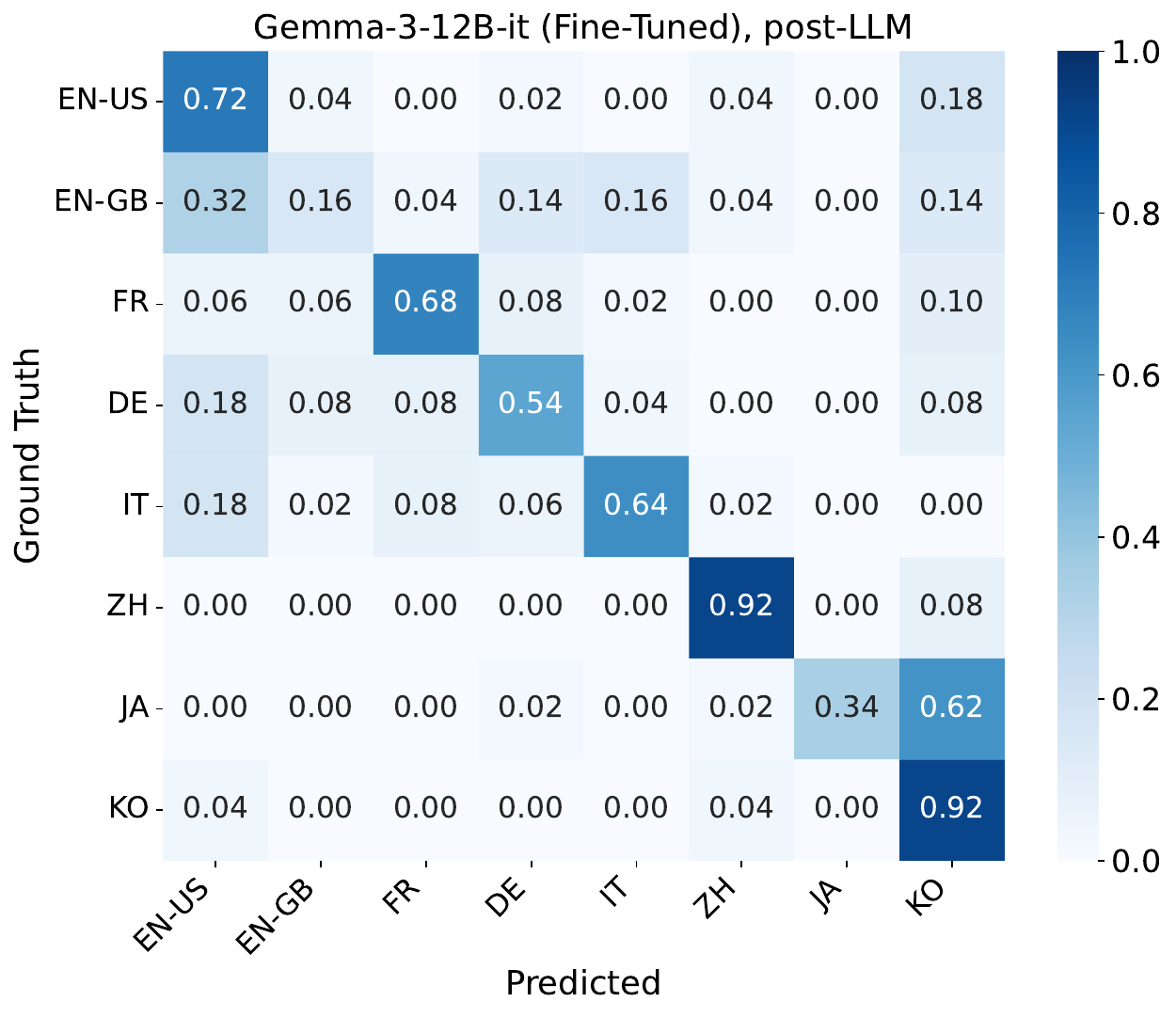}
    \caption{Confusion matrix for Gemma-3-12B-it (fine-tuned, post-LLM).}
    \label{fig:cm_full_gemma_postllm}
\end{figure}

Table \ref{tab:appendix_gemma_detailed_res} presents the full fine-tuning metrics for Gemma 3, and Figures \ref{fig:cm_full_gemma_prenn}--\ref{fig:cm_full_gemma_postllm} visualize the corresponding confusion matrices.

Similar to the Qwen3 model, Gemma shows a significant performance drop for Japanese and Korean in the post-LLM era, with Recall falling to 0.340. Additionally, the model shows a marked decline in detecting British English (en-GB), where the F1-score drops to 0.235. In contrast, Chinese (zh) remains the most detected language. It maintains a high F1-score of 0.885 even in the post-LLM era.

\begin{table*}[h]
\centering
\small
\setlength{\tabcolsep}{4pt} 
\begin{tabular}{l|ccc|ccc|ccc}
\specialrule{1.5pt}{1pt}{2pt}
\textbf{Language} & \multicolumn{3}{c|}{\textbf{Precision}} & \multicolumn{3}{c|}{\textbf{Recall}} & \multicolumn{3}{c}{\textbf{F1-score}} \\
 & pre-NN & pre-LLM & post-LLM & pre-NN & pre-LLM & post-LLM & pre-NN & pre-LLM & post-LLM \\ \hline
en-US & 0.643 & 0.571 & 0.480 & 0.720 & 0.480 & 0.720 & \textbf{0.679} & 0.522 & 0.576 \\
en-GB & 0.619 & 0.790 & 0.444 & 0.520 & 0.300 & 0.160 & \textbf{0.565} & 0.435 & 0.235 \\
fr & 0.674 & 0.630 & 0.773 & 0.660 & 0.680 & 0.680 & 0.667 & 0.654 & \textbf{0.723} \\
de & 0.708 & 0.605 & 0.628 & 0.680 & 0.520 & 0.540 & \textbf{0.694} & 0.559 & 0.581 \\
it & 0.796 & 0.672 & 0.744 & 0.780 & 0.820 & 0.640 & \textbf{ 0.788} & 0.739 & 0.688 \\
zh & 0.738 & 0.656 & 0.852 & 0.900 & 0.840 & 0.920 & 0.812 & 0.737 & \textbf{0.885} \\
ja & 0.736 & 0.683 & 1.000 & 0.780 & 0.560 & 0.340 & \textbf{0.757} & 0.615 & 0.508 \\
ko & 0.833 & 0.540 & 0.434 & 0.700 & 0.820 & 0.920 & \textbf{0.761} & 0.651 & 0.590 \\ 
\specialrule{1.5pt}{1pt}{2pt}
\end{tabular}

\caption{Detailed per-class performance metrics for Gemma-3-12B-it (fine-tuned).}

\label{tab:appendix_gemma_detailed_res}
\end{table*}

% --- Gemma Full: pre-NN ---

% --- Gemma Full: pre-LLM ---

% --- Gemma Full: post-LLM ---

\section{Statistical Significance Testing}
\label{sec:appendix_stats_testing}
We performed Fisher's exact test \cite{upton1992fisher} to assess the statistical significance of the performance differences reported in Table \ref{tab:main_results}. The significance level was set to $\alpha = 0.05$. Tables \ref{tab:fisher_fewshot} and \ref{tab:fisher_finetuned} summarize the results for the few-shot and fine-tuned settings respectively.

\begin{table}[H]
\centering
\small
\resizebox{\columnwidth}{!}{
\begin{tabular}{l|cc|cc}
\specialrule{1.5pt}{1pt}{2pt}
 & \multicolumn{2}{c|}{Qwen3-14B} & \multicolumn{2}{c}{Gemma-3-12B-it} \\
Comparison & p-value & Sig. & p-value & Sig. \\ \midrule
pre-NN vs post-LLM & $<$0.0001 & Yes & 0.0022 & Yes \\
pre-NN vs pre-LLM  & $<$0.0001 & Yes & 0.1568 & No \\
pre-LLM vs post-LLM & 0.2127 & No & 0.0272 & Yes \\
\specialrule{1.5pt}{1pt}{2pt}
\end{tabular}}
\caption{Fisher's exact test results for few-shot setting.}
\label{tab:fisher_fewshot}
\end{table}

\begin{table}[H]
\centering
\small
\resizebox{\columnwidth}{!}{
\begin{tabular}{l|cc|cc}
\specialrule{1.5pt}{1pt}{2pt}
 & \multicolumn{2}{c|}{Qwen3-14B} & \multicolumn{2}{c}{Gemma-3-12B-it} \\
Comparison & p-value & Sig. & p-value & Sig. \\ \midrule
pre-NN vs post-LLM  & 0.0049 & Yes & 0.0002 & Yes \\
pre-NN vs pre-LLM   & 0.0218 & Yes & 0.0083 & Yes \\
pre-LLM vs post-LLM & 0.6583 & No  & 0.3105 & No  \\
\specialrule{1.5pt}{1pt}{2pt}
\end{tabular}}
\caption{Fisher's exact test results for fine-tuned setting.}
\label{tab:fisher_finetuned}
\end{table}

For the fine-tuned models, the accuracy drop from pre-NN to post-LLM is significant for both Qwen3 (p $=$ 0.0049) and Gemma 3 (p $=$ 0.0002). A similar drop is also observed as significant between the pre-NN and pre-LLM for both models. However, the difference between pre-LLM and post-LLM is not significant for either model. This suggests that the most significant shift in writing style occurred with the introduction of neural machine translation rather than LLMs specifically. For the few-shot setting, Qwen3 follows the same pattern as the fine-tuned results. Gemma 3 shows a different pattern, where the pre-LLM to post-LLM drop is significant but the pre-NN to pre-LLM drop is not. This may reflect the generally unstable performance of the few-shot approach.

\section{LLM Rewriting Experiment}
\label{app:rewrite_experiment}

\subsection{Rewriting Prompt}
\label{app:rewrite_prompt}
We experiment with three rewriting prompts of increasing intervention strength:
\begin{itemize}
    \item \textbf{Prompt 1:} \textit{``Please proofread the following academic abstract and title. Preserve the original meaning and technical content. Do not include any comments, explanations, or notes about the changes made.''}
    \item \textbf{Prompt 2:} \textit{``Please rewrite the following academic abstract and title to meet the standard of a high-quality academic publication. Preserve the original meaning and technical content. Do not include any comments, explanations, or notes about the changes made.''}
    \item \textbf{Prompt 3:} \textit{``Please check and fix the grammar, spelling, and phrasing of the following academic abstract and title. Preserve the original meaning and technical content. Do not include any comments, explanations, or notes about the changes made.''}
\end{itemize}

\subsection{Results}
\label{app:rewrite_results}
Tables~\ref{tab:app_rewrite_gemma_qwen}, \ref{tab:app_rewrite_qwen_gemma}, and~\ref{tab:app_rewrite_gpt_gemma} present the full per-language precision, recall, and F1-scores before and after rewriting for all rewriter--classifier combinations across all eras. These complement the averaged summary in Table~\ref{tab:rewrite_macro} of the main paper. The after scores ($\bar{\text{Aft.}}$) are computed from precision and recall averaged across the three rewriting prompts described in Appendix~\ref{app:rewrite_prompt}, with F1 recalculated from the averaged values.

% === Rewriter: Gemma-3-12B-it | Classifier: Qwen3-14B ===
\begin{table*}[t]
\centering
\small
\begin{tabular}{l|cc|cc|cc|cc|cc|cc}
%\setlength{\tabcolsep}{4pt}
%\begin{tabular}{@{ }l|cc|cc|cc|cc|cc|cc@{ }}
\specialrule{1.5pt}{1pt}{2pt}
& \multicolumn{4}{c|}{\textbf{pre-NN}} & \multicolumn{4}{c|}{\textbf{pre-LLM}} & \multicolumn{4}{c}{\textbf{post-LLM}} \\
& \multicolumn{2}{c|}{P} & \multicolumn{2}{c|}{R} & \multicolumn{2}{c|}{P} & \multicolumn{2}{c|}{R} & \multicolumn{2}{c|}{P} & \multicolumn{2}{c}{R} \\
\textbf{Lang.} & Bef. & Aft. & Bef. & Aft. & Bef. & Aft. & Bef. & Aft. & Bef. & Aft. & Bef. & Aft. \\
\midrule
en-US & 0.603 & 0.226 & 0.700 & 0.827 & 0.569 & 0.229 & 0.500 & 0.807 & 0.471 & 0.424 & 0.800 & 0.527 \\
en-GB & 0.651 & 0.259 & 0.560 & 0.140 & 0.696 & 0.307 & 0.320 & 0.093 & 0.737 & 0.378 & 0.280 & 0.060 \\
fr    & 0.781 & 0.584 & 0.640 & 0.147 & 0.720 & 0.872 & 0.720 & 0.133 & 0.812 & 0.721 & 0.600 & 0.360 \\
de    & 0.673 & 0.393 & 0.700 & 0.293 & 0.630 & 0.425 & 0.580 & 0.267 & 0.625 & 0.610 & 0.600 & 0.340 \\
it    & 0.695 & 0.625 & 0.820 & 0.260 & 0.661 & 0.810 & 0.820 & 0.307 & 0.781 & 0.673 & 0.640 & 0.373 \\
zh    & 0.836 & 0.933 & 0.920 & 0.433 & 0.759 & 0.863 & 0.880 & 0.333 & 0.878 & 0.818 & 0.860 & 0.600 \\
ja    & 0.800 & 0.626 & 0.720 & 0.327 & 0.808 & 0.702 & 0.420 & 0.287 & 1.000 & 0.845 & 0.300 & 0.187 \\
ko    & 0.809 & 0.472 & 0.760 & 0.613 & 0.524 & 0.318 & 0.880 & 0.667 & 0.462 & 0.242 & 0.980 & 0.960 \\
\hline
Avg.  & 0.731 & 0.515 & 0.727 & 0.380 & 0.671 & 0.566 & 0.640 & 0.362 & 0.721 & 0.589 & 0.633 & 0.426 \\
\specialrule{1.5pt}{1pt}{2pt}
\end{tabular}
\caption{Per-language precision (P) and recall (R) before and after rewriting by
Gemma-3-12B-it, evaluated by Qwen3-14B. Aft.\ denotes values averaged across three
rewriting prompts.}
\label{tab:app_rewrite_gemma_qwen}
\end{table*}

% === Rewriter: Qwen3-14B | Classifier: Gemma-3-12B-it ===
\begin{table*}[t]
\centering
\small
\begin{tabular}{l|cc|cc|cc|cc|cc|cc}
%\setlength{\tabcolsep}{4pt}
%\begin{tabular}{@{ }l|cc|cc|cc|cc|cc|cc@{ }}
\specialrule{1.5pt}{1pt}{2pt}
& \multicolumn{4}{c|}{\textbf{pre-NN}} & \multicolumn{4}{c|}{\textbf{pre-LLM}} & \multicolumn{4}{c}{\textbf{post-LLM}} \\
& \multicolumn{2}{c|}{P} & \multicolumn{2}{c|}{R} & \multicolumn{2}{c|}{P} & \multicolumn{2}{c|}{R} & \multicolumn{2}{c|}{P} & \multicolumn{2}{c}{R} \\
\textbf{Lang.} & Bef. & Aft. & Bef. & Aft. & Bef. & Aft. & Bef. & Aft. & Bef. & Aft. & Bef. & Aft. \\
\midrule
en-US & 0.643 & 0.443 & 0.720 & 0.673 & 0.571 & 0.415 & 0.480 & 0.373 & 0.480 & 0.253 & 0.720 & 0.867 \\
en-GB & 0.619 & 0.479 & 0.520 & 0.287 & 0.790 & 0.370 & 0.300 & 0.113 & 0.444 & 0.233 & 0.160 & 0.047 \\
fr    & 0.674 & 0.594 & 0.660 & 0.453 & 0.630 & 0.625 & 0.680 & 0.440 & 0.773 & 0.970 & 0.680 & 0.127 \\
de    & 0.708 & 0.640 & 0.680 & 0.547 & 0.605 & 0.522 & 0.520 & 0.367 & 0.628 & 0.479 & 0.540 & 0.327 \\
it    & 0.796 & 0.689 & 0.780 & 0.513 & 0.672 & 0.680 & 0.820 & 0.607 & 0.744 & 0.882 & 0.640 & 0.200 \\
zh    & 0.738 & 0.787 & 0.900 & 0.673 & 0.656 & 0.707 & 0.840 & 0.533 & 0.852 & 0.953 & 0.920 & 0.467 \\
ja    & 0.736 & 0.750 & 0.780 & 0.667 & 0.683 & 0.683 & 0.560 & 0.393 & 1.000 & 0.958 & 0.340 & 0.127 \\
ko    & 0.833 & 0.480 & 0.700 & 0.780 & 0.540 & 0.292 & 0.820 & 0.880 & 0.434 & 0.316 & 0.920 & 0.833 \\
\hline
Avg.  & 0.718 & 0.608 & 0.718 & 0.574 & 0.643 & 0.537 & 0.628 & 0.463 & 0.669 & 0.630 & 0.615 & 0.374 \\
\specialrule{1.5pt}{1pt}{2pt}
\end{tabular}
\caption{Per-language precision (P) and recall (R) before and after rewriting by
Qwen3-14B, evaluated by Gemma-3-12B-it. Aft.\ denotes values averaged across three
rewriting prompts.}
\label{tab:app_rewrite_qwen_gemma}
\end{table*}

% === Rewriter: GPT-4o | Classifier: Gemma-3-12B-it ===
\begin{table*}[t]
\centering
\small
\begin{tabular}{l|cc|cc|cc|cc|cc|cc}
%\setlength{\tabcolsep}{4pt}
%\begin{tabular}{@{ }l|cc|cc|cc|cc|cc|cc@{ }}
\specialrule{1.5pt}{1pt}{2pt}
& \multicolumn{4}{c|}{\textbf{pre-NN}} & \multicolumn{4}{c|}{\textbf{pre-LLM}} & \multicolumn{4}{c}{\textbf{post-LLM}} \\
& \multicolumn{2}{c|}{P} & \multicolumn{2}{c|}{R} & \multicolumn{2}{c|}{P} & \multicolumn{2}{c|}{R} & \multicolumn{2}{c|}{P} & \multicolumn{2}{c}{R} \\
\textbf{Lang.} & Bef. & Aft. & Bef. & Aft. & Bef. & Aft. & Bef. & Aft. & Bef. & Aft. & Bef. & Aft. \\
\midrule
en-US & 0.643 & 0.376 & 0.720 & 0.640 & 0.571 & 0.381 & 0.480 & 0.413 & 0.480 & 0.467 & 0.720 & 0.513 \\
en-GB & 0.619 & 0.409 & 0.520 & 0.180 & 0.790 & 0.280 & 0.300 & 0.067 & 0.444 & 0.391 & 0.160 & 0.067 \\
fr    & 0.674 & 0.621 & 0.660 & 0.387 & 0.630 & 0.696 & 0.680 & 0.340 & 0.773 & 0.667 & 0.680 & 0.267 \\
de    & 0.708 & 0.610 & 0.680 & 0.513 & 0.605 & 0.485 & 0.520 & 0.353 & 0.628 & 0.590 & 0.540 & 0.327 \\
it    & 0.796 & 0.699 & 0.780 & 0.460 & 0.672 & 0.717 & 0.820 & 0.507 & 0.744 & 0.719 & 0.640 & 0.353 \\
zh    & 0.738 & 0.790 & 0.900 & 0.607 & 0.656 & 0.714 & 0.840 & 0.387 & 0.852 & 0.873 & 0.920 & 0.467 \\
ja    & 0.736 & 0.694 & 0.780 & 0.653 & 0.683 & 0.665 & 0.560 & 0.360 & 1.000 & 0.967 & 0.340 & 0.247 \\
ko    & 0.833 & 0.425 & 0.700 & 0.840 & 0.540 & 0.255 & 0.820 & 0.913 & 0.434 & 0.216 & 0.920 & 0.960 \\
\hline
Avg.  & 0.718 & 0.578 & 0.718 & 0.535 & 0.643 & 0.524 & 0.628 & 0.418 & 0.669 & 0.611 & 0.615 & 0.400 \\
\specialrule{1.5pt}{1pt}{2pt}
\end{tabular}
\caption{Per-language precision (P) and recall (R) before and after rewriting by
GPT-4o, evaluated by Gemma-3-12B-it. Aft.\ denotes values averaged across three
rewriting prompts.}
\label{tab:app_rewrite_gpt_gemma}
\end{table*}
\end{document}